\documentclass{article}

\usepackage{microtype}
\usepackage{graphicx}
\usepackage{booktabs} 

\usepackage{hyperref}

\usepackage[accepted]{icml2020}

\usepackage{graphbox}
\usepackage{amsmath}
\usepackage{amsthm}
\usepackage{mathtools}
\usepackage{graphbox} 
\usepackage{thmtools,thm-restate}
\usepackage{graphbox}
\usepackage{pbox}
\usepackage{algorithm,algorithmic}
\usepackage{caption}
\usepackage{subcaption}
\usepackage{amsfonts}
\newcommand{\cifarMSD}{47.0\%}
\newcommand{\cifarTramer}{40.6\%}
\newcommand{\mnistMSD}{58.4\%}

\newcommand\maxt{{\textsc{Max}}}
\newcommand\avgt{{\textsc{Avg}}}
\newcommand\MSD{{\textsc{MSD}}}

\DeclareMathOperator*{\argmin}{arg\;min}
\DeclareMathOperator*{\argmax}{arg\;max}

\DeclareMathOperator*{\sign}{sign}
\DeclareMathOperator*{\clip}{clip}

\icmltitlerunning{Adversarial Robustness Against the Union of Multiple Perturbation Models}

\begin{document}

\twocolumn[
\icmltitle{Adversarial Robustness Against the Union of Multiple Perturbation Models}




\begin{icmlauthorlist}
\icmlauthor{Pratyush Maini}{iit}
\icmlauthor{Eric Wong}{mld}
\icmlauthor{J. Zico Kolter}{csd,bosch}
\end{icmlauthorlist}

\icmlaffiliation{iit}{Department of Computer Science and Engineering, IIT Delhi, India}
\icmlaffiliation{mld}{Machine Learning Department, Carnegie Mellon University, Pittsburgh, Pennsylvania, USA}
\icmlaffiliation{csd}{Computer Science Department, Carnegie Mellon University, Pittsburgh, Pennsylvania, USA}
\icmlaffiliation{bosch}{Bosch Center for Artificial Intelligence, Pittsburgh, Pennsylvania,
USA}

\icmlcorrespondingauthor{Pratyush Maini}{pratyush.maini@gmail.com}

\icmlkeywords{adversarial examples, adversarial training, robust, perturbation, Machine Learning, ICML}

\vskip 0.3in
]
\printAffiliationsAndNotice{}  

\begin{abstract}
Owing to the susceptibility of deep learning systems to adversarial attacks, there has been a great deal of work in developing (both empirically and certifiably) robust classifiers. While most work has defended against a single type of attack, recent work has looked at defending against multiple perturbation models using simple aggregations of multiple attacks. However, these methods can be difficult to tune, and can easily result in imbalanced degrees of robustness to individual perturbation models, resulting in a sub-optimal worst-case loss over the union. In this work, we develop a natural generalization of the standard PGD-based procedure to incorporate multiple perturbation models into a single attack, by taking the worst-case over all steepest descent directions. This approach has the advantage of directly converging upon a trade-off between different perturbation models which minimizes the worst-case performance over the union. With this approach, we are able to train standard architectures which are simultaneously robust against $\ell_\infty$, $\ell_2$, and $\ell_1$ attacks, outperforming past approaches on the MNIST and CIFAR10 datasets and achieving adversarial accuracy of \cifarMSD{} against the union of ($\ell_\infty$, $\ell_2$, $\ell_1$) perturbations with radius  = (0.03, 0.5, 12) on the latter, improving upon previous approaches which achieve \cifarTramer{} accuracy. 

\end{abstract}

\section{Introduction}
Machine learning algorithms have been shown to be susceptible to \emph{adversarial examples} \citep{szegedy2014intriguing} through the existence of data points which can be adversarially perturbed to be misclassified, but are ``close enough'' to the original example to be imperceptible to the human eye. Methods to generate adversarial examples, or ``attacks", typically rely on gradient information, and most commonly use variations of projected gradient descent (PGD) to maximize the loss within a small perturbation region, usually referred to as the adversary's perturbation model.  
A number of heuristic defenses have been proposed to defend against this phenomenon, e.g. distillation \citep{papernot2016distillation} or logit-pairing \citep{kannan2018logit}. However, as time goes by, the original robustness claims of these defenses 
typically don't hold up to more advanced adversaries or more thorough 
attacks \citep{carlini2017towards, engstrom2018evaluating, mosbach2018logit}. 
One heuristic defense that seems to have survived (to this day) is to use 
\emph{adversarial training} against a PGD adversary \citep{madry2018towards}, and 
remains quite popular due to its simplicity and apparent empirical robustness. The method continues to perform well in empirical benchmarks even when compared to recent work in provable defenses, though it comes with no formal guarantees.

While adversarial training has primarily been used to learn models robust to a single perturbation model, some recent work has looked at empirically defending against multiple perturbation models simultaneously. \citet{schott2018towards} proposed a variational autoencoder based architecture to learn an MNIST classifier which was robust to multiple perturbation models, while \citet{tramer2019adversarial} proposed simple aggregations of different adversaries for adversarial training against multiple perturbation models. 

While these approaches can achieve varying degrees of robustness to the considered adversarial perturbation models, in practice it is quite difficult to achieve an optimal trade-off which minimizes the worst-case error in the union of perturbation models. Rather, these approaches tend to converge to suboptimal local minima, resulting in a model that is highly robust to certain perturbation models while failing to defend against others, and the robust performance can often vary substantially across datasets. This results in poor and unpredictable robust performance against the worst-case attack, and indicates that the optimization procedure actually fails to minimize the worst-case loss in the union of the perturbation models. 

We believe that achieving robustness to multiple perturbations is an essential step towards the eventual objective of universal robustness and our work further motivates research in this area.
In this work, we make three main contributions towards learning models which are adversarially robust to multiple perturbation models. First, we demonstrate the inconsistency of previous approaches across datasets, showing that they converge to suboptimal tradeoffs which may not actually minimize the robust objective of worst-case loss over the combined perturbation model. Second, we propose a modified PGD-based algorithm called ``Multi Steepest Descent'' (\MSD{}) for adversarial training, which naturally incorporates different gradient-based perturbation models into a single unified adversary to directly solve the inner optimization problem of finding the worst-case loss. Third, we show empirically that our approach improves upon past work by finding trade-offs between the perturbation models which significantly improve the worst-case robust performance against multiple perturbation models on both MNIST and CIFAR10. Specifically, on MNIST, our model achieves \mnistMSD{} adversarial accuracy against the union of all three attacks $(\ell_\infty, \ell_2, \ell_1)$ for $\epsilon = (0.3, 2.0, 10)$ respectively, substantially improving upon both the ABS models and also simpler aggregations of multiple adversarial attacks, which at best achieve 42.1\% robust accuracy. Additionally, unlike past work, we also train a CIFAR10 model against the union of all three attacks $(\ell_\infty, \ell_2, \ell_1)$, which achieves 47.0\% adversarial accuracy for $\epsilon = (0.03, 0.5, 12)$ and improves upon the simpler aggregations of multiple attacks which can achieve 40.6\% robust accuracy under this perturbation model. In all cases, we find that our approach is able to consistently reduce the worst-case error under the unified perturbation model. Code for reproducing all the results can be found at:  \href{https://github.com/locuslab/robust_union}{https://github.com/locuslab/robust\_union}.

\section{Related work}
After their original introduction, one of the first widely-considered attacks against deep networks had been the Fast Gradient Sign Method \citep{goodfellow2015explaining}, which showed 
that a single, small step in the direction of the sign of the gradient could sometimes fool 
machine learning classifiers. While this worked to some degree, the Basic Iterative Method \citep{kurakin2017adversarial} 
(now typically referred to as the PGD attack) was significantly more successful at creating adversarial examples, and now lies at the core of many papers. 
Since then, a number of improvements and adaptations have been made to the base PGD algorithm to overcome heuristic defenses and create stronger adversaries. Adversarial attacks were thought to be safe under realistic transformations \citep{lu2017no} until the attack was augmented to be robust to them \citep{pmlr-v80-athalye18b}. Adversarial examples generated using PGD on surrogate models can transfer to black box models \citep{papernote2017practical}. Utilizing core optimization techniques such as momentum 
can greatly improve the attack success rate and transferability, and was the winner of the NIPS 2017 competition on adversarial examples \citep{Dong_2018_CVPR}. \citet{uesato2018adversarial} showed that 
a number of ImageNet defenses were not as robust as originally thought, and \citet{obfuscated-gradients} 
defeated many of the heuristic defenses submitted to ICLR 2018 shortly after the reviewing cycle ended, all with stronger PGD variations. 

Throughout this cycle of attack and defense, some defenses were uncovered that 
remain robust to this day. The aforementioned PGD attack, and the related defense known as adversarial training with a PGD adversary (which incorporates PGD-attacked examples into the training process) has so far remained empirically robust \citep{madry2018towards}. Verification methods to certify robustness properties of networks were developed, utilizing techniques such as SMT solvers \citep{katz2017reluplex}, SDP relaxations \citep{raghunathan2018semi}, and mixed-integer linear programming \citep{tjeng2018evaluating}, the last of which has recently been successfully scaled to reasonably sized networks. 
Other work has folded verification into the training process to create provably robust networks 
\citep{wong2018provable, raghunathan2018certified}, some of which have also been scaled to even larger 
networks \citep{wong2018scaling, mirman2018diff, gowal2018interval}. Although some of these could potentially be extended to apply to multiple perturbations simultaneously, most of these works have focused primarily on defending against and verifying only a \emph{single} type of 
adversarial perturbation at a time. 

Last but most relevant to this work are adversarial defenses that are robust against multiple types of attacks simultaneously. \citet{schott2018towards} used multiple variational autoencoders to construct a complex architecture called analysis by synthesis (ABS) for the MNIST dataset that is not as easily attacked by $\ell_\infty$, $\ell_2$, and $\ell_0$ adversaries. The ABS model has two variations, one which is robust to $\ell_0$ and $\ell_2$ but not $\ell_\infty$ attacks and other which is robust to $\ell_\infty$ and $\ell_0$ but not $\ell_2$ attacks. 
Similarly, \citet{tramer2019adversarial} study the theoretical and empirical trade-offs of adversarial robustness in various settings when defending against aggregations of multiple adversaries, however they find that the $\ell_\infty$ perturbation model interferes with other perturbation models on MNIST ($\ell_1$ and $\ell_2$) and they study a rotation and translation adversary instead of an $\ell_2$ adversary for CIFAR10. \citet{croce2019provable} propose a provable adversarial defense against all $\ell_p$ norms for $p\geq 1$ using a regularization term.
Finally, while not studied as a defense, \citet{kang2019transfer} study the transferability of adversarial robustness between models trained against different perturbation models, while \citet{jordan2019quantifying} study combination attacks with low perceptual distortion.

\section{Overview of adversarial training}
Adversarial training is an approach to learn a classifier which minimizes the worst-case 
loss within some perturbation region (the perturbation model). Specifically, for some network $f_\theta$ parameterized by $\theta$, loss function $\ell$, and training data $\{x_i, y_i\}_{i=1\dots n}$, the robust optimization problem of minimizing the worst-case loss within $\ell_p$ norm-bounded perturbations with radius $\epsilon$ is 
\begin{equation}
    \min_\theta \sum_{i} \max_{\delta \in \Delta_{p,\epsilon}} \ell(f_\theta(x_i + \delta), y_i),
    \label{eq:robust_opt}
\end{equation}
where $\Delta_{p,\epsilon} = \{ \delta : \|\delta\|_p \leq \epsilon \}$ is the $\ell_p$ ball with radius $\epsilon$ centered around the origin. 
To simplify the notation, we will abbreviate $\ell(f_\theta(x + \delta), y) = \ell(x + \delta; \theta)$. 

\subsection{Solving the inner optimization problem}
We first look at solving the inner maximization problem, namely 
\begin{equation}
    \max_{\delta\in\Delta_{p,\epsilon}} \ell(x + \delta; \theta). 
\end{equation}
This is the problem addressed by the ``attackers'' in the space of 
adversarial examples, hoping that the classifier can be tricked by the 
optimal perturbed image, $x + \delta^\star$. 
Typical solutions solve this problem by running a form of projected gradient descent, which iteratively takes steps in the gradient direction to increase the loss followed by a projection step back onto the feasible region, the $\ell_p$ ball. 
Since the gradients at the example points themselves (i.e., $\delta = 0$) are typically too small to make efficient progress, more commonly used is a variation called \emph{projected steepest descent}.

\begin{figure}[t]
\centering
\vskip 0 in
  \centering
  \includegraphics[width=0.5\columnwidth]{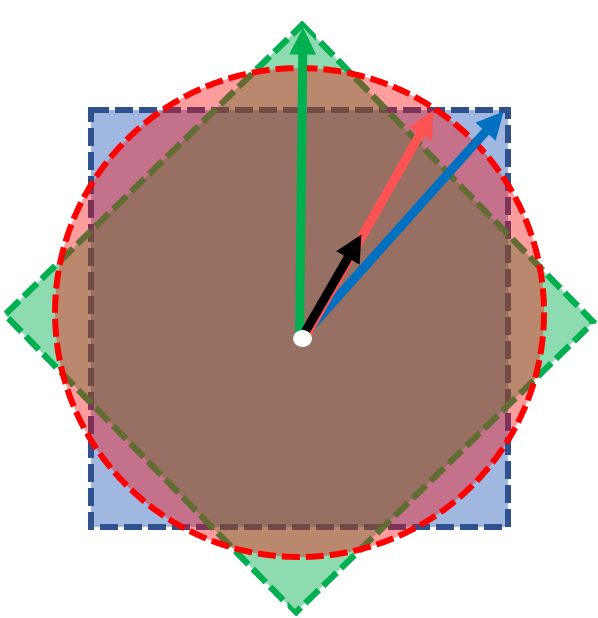}
\caption{A depiction of the steepest descent directions for $\ell_\infty$, $\ell_2$, and $\ell_1$ norms. The gradient is the black arrow, and the $\alpha$ radius step sizes and their corresponding steepest descent directions $\ell_\infty$, $\ell_2$, and $\ell_1$ are shown in blue, red, and green respectively. 
}
\label{fig:steepest_l2_ex}
\vskip -0.1in
\end{figure}

\paragraph{Steepest descent} 
 For some norm $\|\cdot \|_p$ and step size $\alpha$, the direction of steepest descent on the loss function $\ell$ for a perturbation $\delta$ is 
\begin{equation}
    v_p(\delta) = \argmax_{\|v\|_p \leq \alpha} v^T\nabla \ell(x + \delta; \theta).
    \label{eq:steepest_descent}
\end{equation}
Then, instead of taking gradient steps, steepest descent uses the following iteration 
\begin{equation}
    \delta^{(t+1)} = \delta^{(t)} + v_p(\delta^{(t)}).
\end{equation}

In practice, the norm used in steepest descent is typically taken to be the same $\ell_p$ norm used 
to define the perturbation region $\Delta_{p,\epsilon}$. However, depending on the norm used, the direction of steepest descent can be quite different from the actual gradient (Figure \ref{fig:steepest_l2_ex}). Note that a single steepest descent step with respect to the $\ell_\infty$ norm reduces to $v_\infty(x) = \alpha \cdot \sign(\nabla \ell(x + \delta; \theta))$, better known in the adversarial examples literature as the Fast Gradient Sign Method \citep{goodfellow2015explaining}.

\paragraph{Projections}
The second component of projected steepest descent for adversarial examples is 
to project iterates back onto the $\ell_p$ ball around $x$. Specifically, 
projected steepest descent performs the following iteration 
\begin{equation}
    \delta^{(t+1)} = \mathcal P_{\Delta_{p,\epsilon}} \left(\delta^{(t)} + v_p(\delta^{(t)})\right)
\end{equation}
where $\mathcal{P}_{\Delta_{p,\epsilon}}(\delta)$ is the standard projection operator that finds the perturbation $\delta'\in \Delta_{p,\epsilon}$ that is ``closest'' in Euclidean space to the input $\delta$, defined as 
\begin{equation}
    \mathcal P_{\Delta_{p,\epsilon}}(\delta)  = \argmin_{\delta'\in \Delta_{p,\epsilon}} \|\delta - \delta'\|_2^2. 
\end{equation}
Visually, a depiction of this procedure (steepest descent followed by a projection onto the perturbation region) for an $\ell_2$ adversary can be found in Figure \ref{fig:steepest_l2_ex}. If we instead project the steepest descent directions with respect to the $\ell_\infty$ norm onto the $\ell_\infty$ ball of allowable perturbations, the projected steepest descent iteration reduces to 
\begin{equation}
\begin{split}
    \delta^{(t+1)} &= P_{\Delta_{\infty,\epsilon}}(\delta^{(t)} + v_\infty(\delta^{(t)})) \\
    &= \clip_{[-\epsilon, \epsilon]}\left(\delta^{(t)} + \alpha \cdot \sign(\nabla \ell(x + \delta^{(t)}; \theta))\right) 
\end{split}
\end{equation}
where $\clip_{[-\epsilon, +\epsilon]}$ ``clips'' the input to lie within the range $[-\epsilon, \epsilon]$. This is exactly the Basic Iterative Method used in \citet{kurakin2017adversarial}, 
typically referred to in the literature as an $\ell_\infty$ PGD adversary.  

\subsection{Solving the outer optimization problem}
We next look at how to solve the outer optimization problem, or the problem of learning the weights $\theta$ that minimize the loss of our classifier. While many approaches have been proposed in the literature, we will focus on a heuristic called adversarial training, which has generally worked well in practice. 

\paragraph{Adversarial training}
Although solving the min-max optimization problem may seem daunting, a classical result known as Danskin's theorem \citep{danskin1967theory} says that the gradient of a maximization problem is equal to the gradient of the objective evaluated at the optimum. For learning models that minimize the robust 
optimization problem from Equation \eqref{eq:robust_opt}, this means that 
\begin{equation}
    \nabla_\theta \left(\sum_{i} \max_{\delta \in \Delta_{p,\epsilon}} \ell(x_i + \delta; \theta)\right) =\sum_{i} \nabla_\theta \ell(x_i + \delta^*(x_i); \theta)
\end{equation}
where $\delta^*(x_i) = \argmax_{\delta \in \Delta_{p,\epsilon}} \ell(x_i + \delta; \theta)$. 
In other words, this means that in order to backpropagate through the robust optimization problem, 
we can solve the inner maximization and backpropagate through the solution. 
Adversarial training does this by empirically maximizing the inner problem with a PGD 
adversary. Note that since the inner problem is not solved exactly, Danskin's theorem does not strictly apply. However, in practice, adversarial training does seem to provide good empirical robustness, at least when evaluated against the $\ell_p$ perturbation model it was trained against. 

\section{Adversarial training for multiple perturbation models}
We can now consider the core of this work, adversarial training procedures against multiple perturbation models. 
More formally, let $\mathcal S$ represent a set of perturbation models, such that $p \in \mathcal S$ corresponds to the $\ell_p$ perturbation model $\Delta_{p,\epsilon}$, and 
let $\Delta_{\mathcal S} = \bigcup_{p \in \mathcal S}\Delta_{p,\epsilon}$ be the union of all 
perturbation models in $\mathcal S$. Note that the $\epsilon$ chosen for each ball is \emph{not} typically the same, but we still use the same notation $\epsilon$ for simplicity, since the context will always make clear which $\ell_p$-ball we are talking about.
Then, the generalization of the robust optimization problem in Equation \eqref{eq:robust_opt} to multiple perturbation models is   
\begin{equation}
    \min_\theta \sum_{i}\max_{\delta \in \Delta_{\mathcal S}} \ell(x_i + \delta; \theta). 
    \label{eq:robust_opt_multiple}
\end{equation}
The key difference is in the inner maximization, where the worst-case adversarial loss is now taken over \emph{multiple} $\ell_p$ perturbation models. In order to perform adversarial training, using the same motivational idea from Danskin's theorem, we can backpropagate through the inner maximization by first finding (empirically) the optimal perturbation,  
\begin{equation}
    \delta^* = \argmax_{\delta \in \Delta_\mathcal{S}} \ell(x + \delta; \theta). 
    \label{eq:optimal}
\end{equation}
To find the optimal perturbation over the union of perturbation models, we begin by discussing simple  generalizations of standard adversarial training, which will use aggregations of PGD solutions for individual adversaries to approximately solve the inner maximization over multiple adversaries. The computational complexity of these approaches are a constant factor times than the complexity of standard adversarial training, where the constant is equal to the number of adversaries. We will focus the exposition primarily on adversarial training based approaches as these are most related to our proposed method, and we refer the reader to \citet{schott2018towards} for more detail on the analysis by synthesis approach. 

\subsection{Simple combinations of multiple perturbations}
First, we study two simple approaches to generalizing adversarial training to multiple perturbation models, which can learn robust models and do not rely on complicated architectures. While these methods work to some degree, we later find empirically that these methods do not necessarily minimize the worst-case performance, can converge to unexpected tradeoffs between multiple perturbation models, and can have varying dataset-dependent performance.  

\paragraph{\maxt{}: Worst-case perturbation}
One way to generalize adversarial training to multiple perturbation models is to 
use each perturbation model independently, and train on 
the adversarial perturbation that achieved the maximum loss. Specifically, for each adversary $p \in \mathcal S$, we 
solve the innermost maximization with an $\ell_p$ PGD adversary to get an approximate worst-case 
perturbation $\delta_p$, 
\begin{equation}
    \delta_p =  \argmax_{\delta \in \Delta_{p,\epsilon}} \ell(x + \delta; \theta), 
    \label{eq:subproblem}
\end{equation}
and then approximate the maximum over all adversaries as 
\begin{equation}
    \delta^* \approx \argmax_{\delta_p} \ell(x + \delta_p; \theta).
\end{equation}
When $|\mathcal S| = 1$, then this reduces to standard adversarial training. Note that if each PGD adversary solved their subproblem from Equation \eqref{eq:subproblem} exactly, then this is the optimal perturbation $\delta^\star$. This method corresponds to the ``max'' strategy from \citet{tramer2019adversarial}. 

\paragraph{\avgt{}: Augmentation of all perturbations}
Another way to generalize adversarial training is to train on all the adversarial perturbations 
for all $p \in \mathcal S$ to form a larger adversarial dataset. Specifically, instead of solving 
the robust problem for multiple adversaries in 
Equation \eqref{eq:robust_opt_multiple}, we instead solve 
\begin{equation}
    \min_\theta \sum_{i}\sum_{p \in \mathcal S} \max_{\delta \in \Delta_{p,\epsilon}} \ell(x_i + \delta; \theta) 
    \label{eq:pgd_augmentation}
\end{equation}
by using individual $\ell_p$ PGD adversaries to approximate the inner maximization for each perturbation model. This reduces to standard adversarial training when $|\mathcal S| = 1$ and corresponds to the ``avg'' strategy from \citet{tramer2019adversarial}. 

While these methods work to some degree, (which is shown later in Section \ref{sec:results}), both of these approaches solve the inner maximization problem independently for each adversary. Consequently, each individual PGD adversary is myopic to its own perturbation model and does not take advantage of the fact that the perturbation region is enlarged by other perturbation models. To leverage the full information provided by the union of perturbation regions, we propose a modification to standard adversarial training, which combines information from all considered perturbation models into a single PGD adversary that is potentially stronger than the combination of independent adversaries.  

\begin{algorithm}[tb]
  \caption{Multi steepest descent for learning classifiers that are simultaneously robust to $\ell_p$ 
  attacks for $p \in \mathcal S$}    
    \begin{algorithmic}
  \STATE {\bfseries Input:} classifier $f_\theta$,  data $x$, labels $y$
  \STATE{\bfseries Parameters:} $\epsilon_p, \alpha_p$ for $p\in \mathcal S$, maximum iterations $T$, loss function $\ell$
  \STATE $\delta^{(0)} = 0$
  \FOR{$t=0\dots T-1$}
    \FOR {$p \in \mathcal S$}
    \STATE $\delta^{(t+1)}_p = P_{\Delta_{p, \epsilon}}(\delta^{(t)} + v_p(\delta^{(t)}))$
    \ENDFOR
    \STATE $\delta^{(t+1)} = \argmax_{\delta_p^{(t+1)}} \ell(f_\theta(x + \delta_p^{(t+1)}), y)$
  \ENDFOR
  \STATE return $\delta^{(T)}$
\end{algorithmic}
\label{alg:proposed}
\end{algorithm}

\subsection{Multi Steepest Descent}
To create a PGD adversary with full knowledge of the perturbation region, 
we propose an algorithm that incorporates the different perturbation models within each step of projected steepest descent. Rather than generating adversarial examples for each perturbation model with separate PGD adversaries, the core idea is to create a single adversarial perturbation by simultaneously maximizing the worst-case loss over all perturbation models at each projected steepest descent step. We call our method \emph{multi steepest descent} (\MSD{}), 
which can be summarized as the following iteration: 
\begin{equation}
\begin{split}
    \delta^{(t+1)}_p &= P_{\Delta_{p, \epsilon}}(\delta^{(t)} + v_p(\delta^{(t)})) \;\; \text{for} \;\; p \in \mathcal S\\
    \delta^{(t+1)} &= \argmax_{\delta_p^{(t+1)}} \ell(x + \delta^{(t+1)}_p)
\end{split}
\label{eq:ours}
\end{equation}
The key difference here is that at each iteration of \MSD{}, we choose a projected steepest descent direction 
that maximizes the loss over all attack models $p \in \mathcal S$, whereas standard adversarial 
training and the simpler approaches use comparatively myopic PGD  subroutines that only use one 
perturbation model at a time. The full algorithm is in 
Algorithm \ref{alg:proposed}, and can be used as a drop in replacement for standard PGD adversaries to learn robust classifiers with adversarial training. 
We direct the reader to Appendix \ref{app:steep} for a complete description of steepest descent directions and projection operators for $\ell_\infty$, $\ell_2$, and $\ell_1$ norms.\footnote{The pure $\ell_1$ steepest descent step is inefficient since it only updates one coordinate at a time. It can be improved by taking steps on multiple coordinates, similar to that used in \citet{tramer2019adversarial}, and is also explained in Appendix \ref{app:steep}.}

\section{Results}
\label{sec:results}
In this section, we present experimental results on using generalizations of adversarial training to achieve simultaneous robustness to $\ell_\infty$, $\ell_2$, and $\ell_1$ perturbations on the MNIST and CIFAR10 datasets. 
Our primary goal is to show that adversarial training can be used to directly minimize the worst-case loss over the union of perturbation models to achieve competitive results by avoiding any trade-off that biases one particular perturbation model at the cost of the others. 
Our results improve upon the state-of-the-art in three key ways. First, we can continue to use simple, standard architectures for image classifiers, without relying on complex architectures or input binarization as done by \citet{schott2018towards}. Second, our method is able to learn a single model (on both MNIST and CIFAR10) which optimizes the worst-case performance over the union of all three perturbation models, whereas previous approaches are only robust against two at a time, or have performance which is dataset dependent. Finally, we provide the first CIFAR10 model trained to be simultaneously robust against $\ell_\infty$, $\ell_2$, and $\ell_1$ adversaries, in comparison to previous work which trained a model robust to $\ell_\infty$, $\ell_1$, and rotation/translation attacks \citep{tramer2019adversarial}. 

We train models using \MSD{}, \maxt{} and \avgt{} approaches for both MNIST and CIFAR10 datasets. We additionally train models against individual PGD adversaries to measure the changes and tradeoffs in universal robustness. Since the analysis by synthesis model is not scalable, we do not include it in our experimentation for CIFAR10. We perform an extensive evaluation of these models with a broad suite of both gradient and non-gradient based attacks using Foolbox\footnote{https://github.com/bethgelab/foolbox \citep{rauber2017foolbox}} (the same attacks used by \citet{schott2018towards}), and also incorporate all the PGD-based adversaries discussed in this paper. All aggregate statistics that combine multiple attacks compute the worst-case error rate over all attacks for \emph{each} example, in order to reflect the worst-case loss over the combined perturbation model. 

Summaries of these results at specific thresholds can be found in Tables \ref{table:mnist} and \ref{table:cifar10}, where B-ABS and ABS refer to binarized and non-binarized versions of the analysis by synthesis models from \citet{schott2018towards}, $P_p$ refers to a model trained against a PGD adversary with respect to the $p$-norm, \maxt{} and \avgt{} refer to models trained using the worst-case and data augmentation generalizations of adversarial training, and MSD refers to models trained using multi steepest descent. Full tables containing the complete breakdown of these numbers over all individual attacks used in the evaluation are in Appendix \ref{app:evaluation}. We report the results against individual attacks and perturbation models for completeness, however we note that the original goal and motivation of all these algorithms is to minimize the robust optimization objective from Equation \eqref{eq:robust_opt_multiple}. While there may be different implicit tradeoffs between individual perturbation models that can be difficult to compare, the robust optimization objective, or the performance against the union of \emph{all} attacks, provides a single common metric that all approaches are optimizing.  


\subsection{Experimental setup}
\label{sec:Arch}
\paragraph{Architectures and hyperparameters}
For MNIST, we use a four layer convolutional network with two convolutional layers consisting of 32 and 64 $5 \times 5$ filters and 2 units of padding, followed by a fully connected layer with 1024 hidden units, where both convolutional layers are followed by $2 \times 2$ Max Pooling layers and ReLU activations (this is the same architecture used by \citet{madry2018towards}). This is in contrast to past work on MNIST, which relied on per-class variational autoencoders to achieve robustness against multiple perturbation models \citep{schott2018towards}, which was also not easily scalable to larger datasets. Since our methods have the same computational complexity as standard adversarial training, they also easily apply to standard CIFAR10 architectures, and in this paper we use the well known pre-activation version of the ResNet18 architecture consisting of nine residual units with two convolutional layers each \citep{he2016identity}. 

A complete description of the hyperparameters used is in Appendix \ref{app:setup}. All reported $\epsilon$ are for images scaled to be between the range $[0,1]$. All experiments were run on modest amounts of GPU hardware (e.g. a single 1080ti). 



\paragraph{Attacks used for evaluation}
To evaluate the model, we incorporate the attacks from \citet{schott2018towards} along with our PGD based adversaries, and provide a short description of the same here. Note that we exclude attacks based on gradient estimation, since the gradient for the standard architectures used here are readily available. 

For $\ell_\infty$ attacks, although we find the $\ell_\infty$ PGD adversary to be quite effective, for completeness, we additionally use the Foolbox implementations of Fast Gradient Sign Method \citep{goodfellow2015explaining}, PGD attack \citep{madry2018towards}, and Momentum Iterative Method \citep{Dong_2018_CVPR}. 

For $\ell_2$ attacks, in addition to the $\ell_2$ PGD adversary, we use the Foolbox implementations of the same PGD adversary, the Gaussian noise attack \citep{rauber2017foolbox}, the boundary attack \citep{brendel2017decision}, DeepFool \citep{moosavi2016deepfool}, the pointwise attack \citep{schott2018towards}, DDN based attack \citep{rony2018ddn}, and C\&W attack \citep{carlini2017towards}. 

For $\ell_1$ attacks, we use both the $\ell_1$ PGD adversary as well as additional Foolbox implementations of $\ell_0$ attacks at the same radius, namely the salt \& pepper attack \citep{rauber2017foolbox} and the pointwise attack \citep{schott2018towards}. Note that an $\ell_1$ adversary with radius $\epsilon$ is strictly stronger than an $\ell_0$ adversary with the same radius, and so we choose to explicitly defend against $\ell_1$ perturbations instead of the $\ell_0$ perturbations considered by \citet{schott2018towards}. 

We make \textbf{10 random restarts} for each of the results mentioned hereon for both MNIST and CIFAR10  \footnotemark\addtocounter{footnote}{0}. 
We encourage future work in this area to incorporate the same, since the success of all attacks, specially decision based or gradient free ones, is observed to increase significantly over restarts.

\footnotetext{\label{fn:restarts}All attacks were run on a subset of the first 1000 test examples with 10 random restarts, with the exception of Boundary Attack, which by default makes 25 trials per iteration and DDN based Attack which does not benefit from the same owing to a deterministic initialization of $\delta$.}

\begin{table*}
  \caption{Summary of adversarial accuracy results for MNIST (higher is better)}
  \label{table:mnist}
  \centering
  \begin{tabular}{l|rrrrrrrr}
    \hline
                          & $P_\infty$ & $P_2$ & $P_1$ & B-ABS\footnotemark\addtocounter{footnote}{-1} & ABS\footnotemark 
                          & \maxt{} & \avgt{} & \MSD{} \\
    \hline 
    Clean Accuracy                          & 99.1\% & 99.2\% & 99.3\% & 99\% & 99\% & 98.6\% & 99.1\% & 98.3\% \\
    \hline
    $\ell_\infty$ attacks $(\epsilon=0.3)$ & 90.3\% & 0.4\% & 0.0\% & 77\% &  8\% & 51.0\% & 65.2\% & 62.7\%\\
    $\ell_2$ attacks $(\epsilon=2.0)$      & 13.6\% & 69.2\% &38.5\% & 39\% & 80\% & 61.9\% & 60.1\% & 67.9\%\\
    $\ell_1$ attacks $(\epsilon=10)$       & 4.2\% & 43.4\% & 70.0\% & 82\% & 78\% & 52.6\% & 39.2\% & 65.0\%\\
    \hline
    All Attacks                            & 3.7\% & 0.4\% & 0.0\% & 39\% &  8\% & 42.1\% & 34.9\% & \textbf{58.4}\% \\
    \hline
  \end{tabular}
\end{table*}

\begin{figure*}[t]
  \centering
  \includegraphics[width=0.3\textwidth]{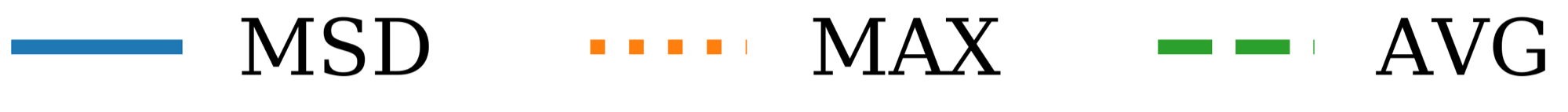} \\
  \vspace{0.05in}
  \includegraphics[width=0.23\textwidth]{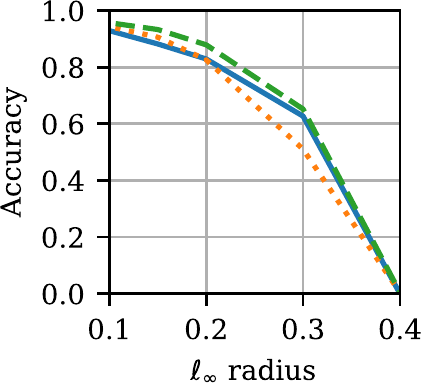}
  \includegraphics[width=0.2\textwidth]{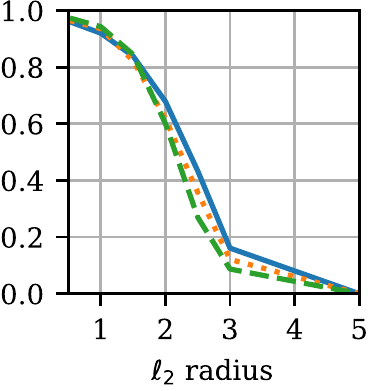}
  \includegraphics[width=0.205\textwidth]{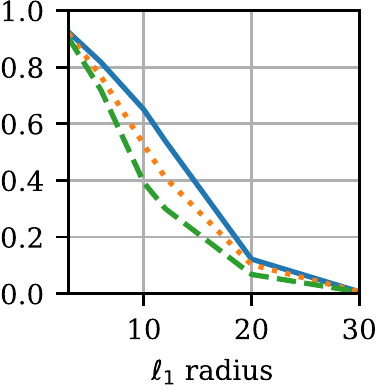}
  \caption{Robustness curves showing the adversarial accuracy for the MNIST model trained with \MSD{}, \avgt{}, \maxt{} against $\ell_\infty$ (left), $\ell_2$ (middle), and $\ell_1$ (right) perturbation models over a range of epsilon.}
  \label{fig:mnist_curve}
\end{figure*}

\subsection{MNIST}
We first present results on the MNIST dataset, which are summarized in Table \ref{table:mnist} (a more detailed breakdown over each individual attack is in Appendix \ref{app:mnist}). Complete robustness curves over a range of epsilons over each perturbation model can be found in Figure \ref{fig:mnist_curve}. Although we reproduce the simpler approaches here, a more detailed discussion of how these results compare with those presented by \citet{tramer2019adversarial} can be found in Appendix \ref{app:tramer}.


\paragraph{Suboptimal trade-offs}
While considered an ``easy'' dataset, we first note that most of the previous approaches for multiple perturbation models on MNIST are only able to defend against two out of three perturbation models at a time, resulting in a suboptimal trade-off between different perturbation models which has poor overall performance against the worst-case attack in the combined perturbation model. Despite relying on a significantly more complex architecture, the B-ABS model is weak against $\ell_2$ attacks while the ABS model is weak against $\ell_\infty$ attacks. Meanwhile, the \avgt{} model is weak against strong $\ell_1$ decision-based attacks. The \maxt{} and \MSD{} models achieve relatively better trade-offs, with the \MSD{} model performing the best with a robust accuracy rate of \mnistMSD{} against the union of  $(\ell_\infty$, $\ell_2$, $\ell_1)$ perturbations with radius $\epsilon = (0.3$, $2.0$, $10$), which is over a 15\% improvement in comparison to the \maxt{} model. 

\begin{figure}[t]
  \centering
  \includegraphics[width=0.7\columnwidth]{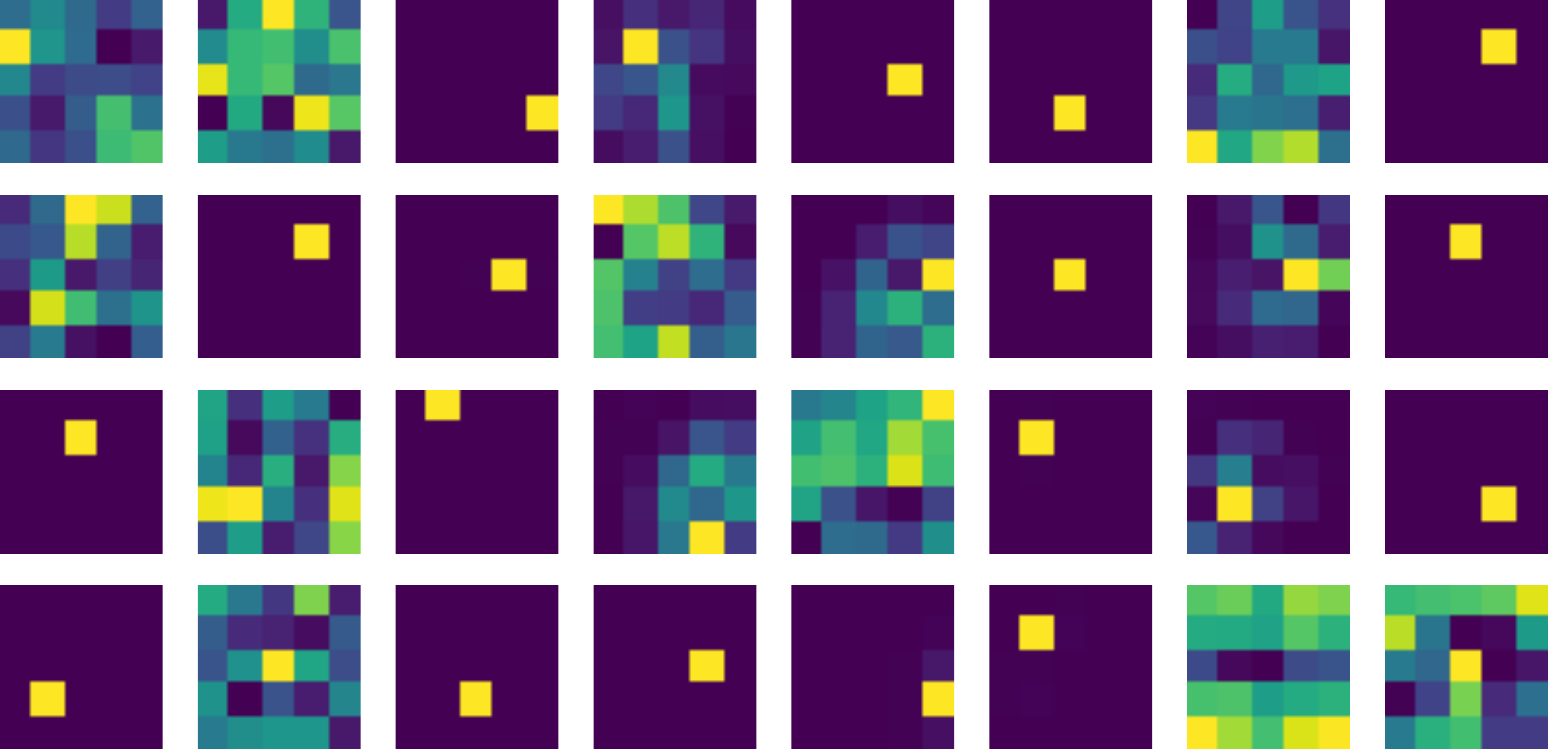}
\\
  \vspace{0.1in}
  \centering
  \includegraphics[width=0.9\columnwidth]{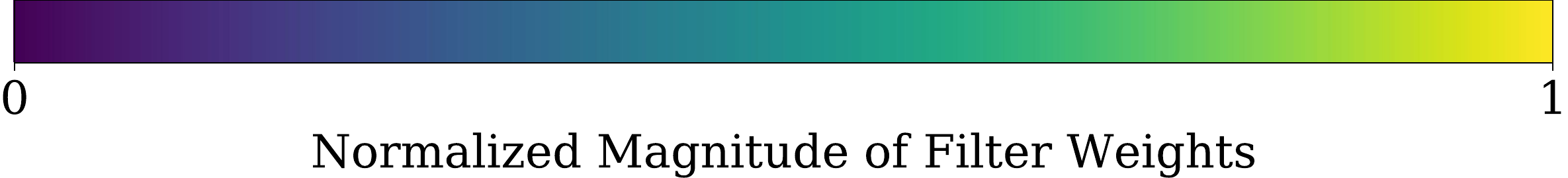}
  \caption{A view of each of the (5x5) learned filters of the first layer of a CNN robust to $\ell_\infty$ attacks. The singular sharp values are characteristic features of models robust to $\ell_\infty$ attacks.}
  \label{fig:linf_filters}
\end{figure}

\paragraph{Gradient Masking in MNIST models} We find that even though models trained via the \maxt{} and \avgt{} approaches provide reasonable robustness against \textit{first-order} attacks (breakdown of attacks in Appendix \ref{app:mnist}), they can be vulnerable to gradient-free attacks like the Pointwise Attack and Boundary Attack. This indicates the presence of masked gradients that prevent \textit{first-order} adversaries from finding the optimal steepest descent direction \citep{obfuscated-gradients}, similar to how $\ell_\infty$ trained models are weak against decision-based attacks in other norms as also observed by \citet{schott2018towards} and \citet{tramer2019adversarial}. 
\footnotetext{\label{fn:mnist}Results are reported directly from \citet{schott2018towards}, which used epsilon balls of radii (0.3,1.5,12) for ($\ell_\infty$, $\ell_2$, $\ell_0$) adversaries. They used an $\ell_0$ perturbation region of a higher radius and evaluated against $\ell_0$ attacks. So the reported number is a near estimate of the $\ell_1$ adversarial accuracy. They used an $\ell_2$ perturbation model of a lower radius = 1.5. Further, they do not perform attack restarts and the adversarial accuracy against all attacks is an upper bound based on the reported accuracies for individual perturbation models. Finally, all ABS results were computed using numerical gradient estimation, since gradients are not readily available.}
We analyze the learned weights of the first layer filters of the CNN models trained on the MNIST, and observe a strong correlation of the presence of thresholding filters (Figure \ref{fig:linf_filters}) with 
the susceptibility to decision-based $\ell_1$ and $\ell_2$ adversaries. Further analysis of the learned filter weights for all the models can be found in Appendix \ref{app:filters}, where we observe that by reducing the number of thresholding filters, the \MSD{} model is able to perform better against decision based adversaries, whereas learning filter patterns similar to that of an $\ell_\infty$ robust model correlates with susceptibility of \maxt{} and \avgt{} training methods to gradient-free adversaries. 

\begin{figure}[t]
  \centering
  \includegraphics[width=0.7\columnwidth]{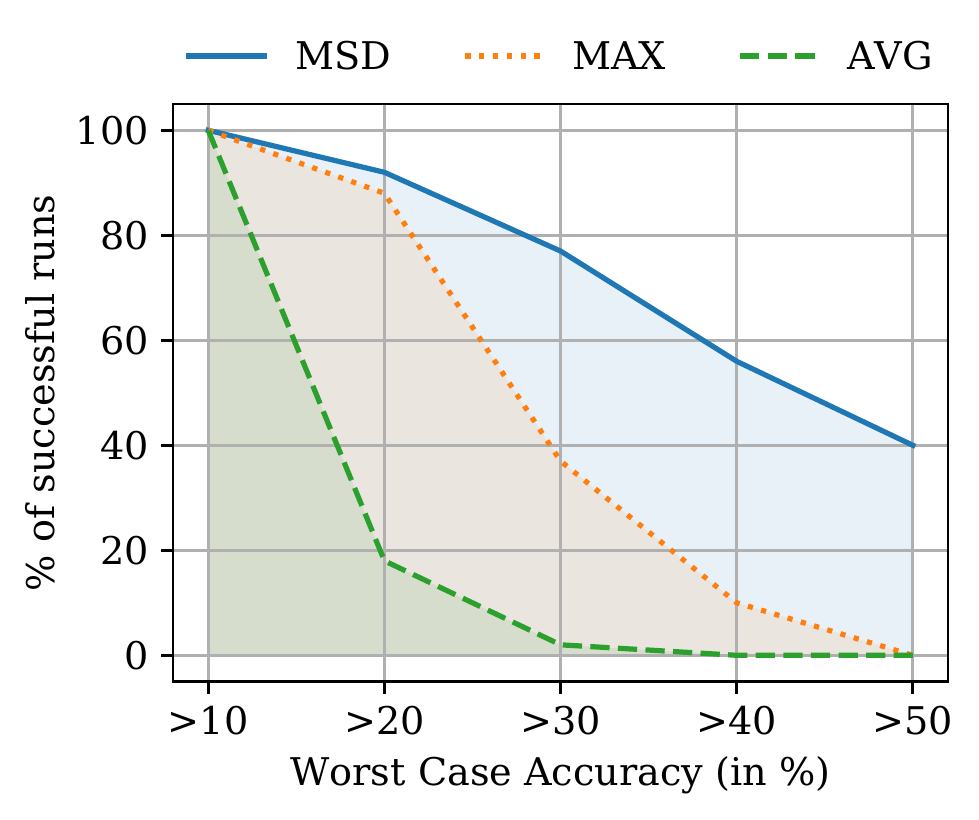}
  \caption{Among all the models trained using the MSD, \maxt{} and \avgt{} methods during our hyperparameter search, we plot the percentage of models for each method that achieve robust accuracies greater than a particular threshold (against the union of $\ell_\infty, \ell_1, \ell_2$ attacks).}
  \label{fig:grid-search}
\end{figure}

\begin{table*}[t]
  \caption{Summary of adversarial accuracy results for CIFAR10 (higher is better)}
  \label{table:cifar10}
  \centering
  \begin{tabular}{l|rrrrrr}
    \hline
                                & $P_\infty$ & $P_2$ & $P_1$ & \maxt{} & \avgt{} & \MSD{} \\
    \hline
    Clean accuracy                           & 83.3\% & 90.2\% & 73.3\% & 81.0\% & 84.6\% & 81.1\% \\
    \hline
    $\ell_\infty$ attacks $(\epsilon=0.03)$& 50.7\% & 28.3\% & 0.2\% & 44.9\% & 42.5\% & 48.0\%\\
    $\ell_2$ attacks $(\epsilon=0.5)$      & 57.3\% & 61.6\% & 0.0\% & 61.7\% & 65.0\% & 64.3\%\\
    $\ell_1$ attacks $(\epsilon=12)$       & 16.0\% & 46.6\% & 7.9\% & 39.4\% & 54.0\% & 53.0\%\\
    \hline
    All attacks                            &15.6\% & 27.5\% & 0.0\% & 34.9\% & 40.6\% & \textbf{47.0}\%\\
    \hline
  \end{tabular}
\end{table*}
\begin{figure*}[t]
  \centering
  \includegraphics[width=0.3\textwidth]{figures/labels.png}
  \\
  \vspace{0.05in}
  \includegraphics[width=0.22\textwidth]{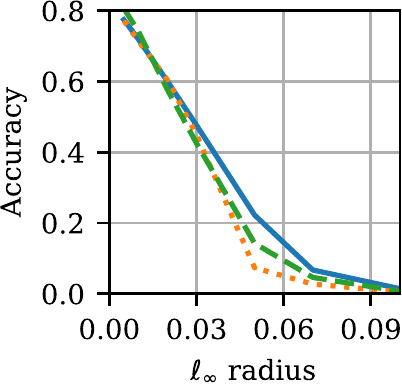}
  \includegraphics[width=0.2\textwidth]{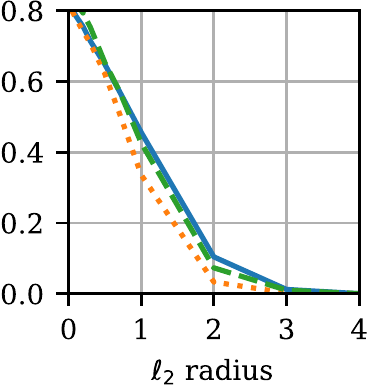}
  \includegraphics[width=0.21\textwidth]{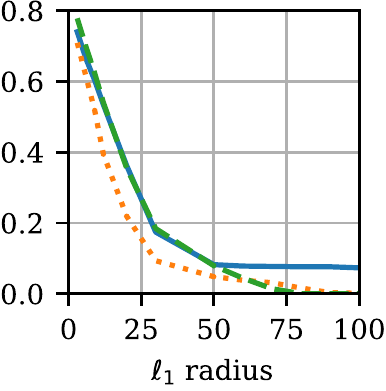}
  \caption{Robustness curves showing the adversarial accuracy for the CIFAR10 model trained with \MSD{}, \avgt{}, \maxt{} against $\ell_\infty$ (left), $\ell_2$ (middle), and $\ell_1$ (right) perturbation models over a range of epsilon.}
  \label{fig:cifar10_curve}
\end{figure*}

\paragraph{Unreliable training of \maxt{} and \avgt{}}
To give the \maxt{} and \avgt{} approaches the best chance at succeeding, we searched over a wide range of hyperparameters (which are described in Appendix \ref{app:hp_adv_training}). However, we frequently observe that these training runs result in masked gradients as described earlier, and are seemingly unable to balance the right trade-off between multiple attacks. In Figure~\ref{fig:grid-search}, we show the sensitivity of different training methods to training time hyperparameter choices. The worst case accuracy is evaluated using the worst case over three gradient based attacks (PGD attacks in $\ell_\infty,\ell_2,\ell_1$ space) and one gradient-free attack (pointwise attack in $\ell_1$ space). The \maxt{} training method achieves greater than 40\% robust accuracy in only 10\% of all the hyperparameter configurations tried. The sensitivity was even higher for the \avgt{} method on the MNIST dataset. Also, note that nearly all models attain greater than 50\% robust accuracy when only attacked by gradient-based adversaries, and the performance drop is largely attributed to the gradient-free attack.

However, MSD is comparatively much easier to tune and achieves greater than 50\% accuracy in around 40\% of the runs. Moreover, we find that MSD offers a natural way to counteract any unwanted bias towards one perturbation type by adjusting the relative step-sizes of individual descent directions, whereas doing the same for the \maxt{} and \avgt{} approaches does not help. 

We note that in order to train the \maxt{} and \avgt{} approaches reasonably well on the MNIST dataset (Table \ref{table:mnist}), we had to set the radius of the $\ell_1$ ball to 12 for \avgt{} and increase the number of PGD $\ell_1$ attack restarts during training for \maxt{}. These methods help make the PGD $\ell_1$ attack relatively \textit{stronger} by changing the perturbation model, and re-aligns the optimal trade-offs when the training process is unable to naturally capture them. We observe that \textit{small starts}, as employed by \citet{tramer2019adversarial} to make their models work better, may have a similar effect of re-aligning the strength of various perturbation models. Rather than ``fixing'' the balance between different perturbation models by changing the individual attacks used for training, MSD is able to achieve the right trade-off by directly balancing them, leading to greater reliability and consistency when compared to the \maxt{} and \avgt{} approaches. 

\subsection{CIFAR10}
Next, we present results on the CIFAR10 dataset, which are summarized in Table \ref{table:cifar10} (a more detailed breakdown over each individual attack is in Appendix \ref{app:cifar10}). Our \MSD{} approach reaches the best performance against the union of attacks, and achieves $47.0\%$ (individually $48.0\%, 64.3\%, 53.0\%$) adversarial accuracy against the union of $(\ell_\infty, \ell_2, \ell_1)$ perturbations of size $\epsilon = (0.03,0.5,12)$. We note that the $P_1$ model trained against an $\ell_1$ PGD adversary is not very robust when evaluated against decision-based attacks, even though it can defend reasonably well against the $\ell_1$ PGD attack in isolation
(Table \ref{table:cifar10_extended} in Appendix \ref{app:cifar10}). Complete robustness curves over a range of epsilons over each perturbation model can be found in Figure \ref{fig:cifar10_curve}. 
The specific heuristic adjustments made to obtain the best-performing \maxt{} and \avgt{} models are detailed in Appendix \ref{app:hp_adv_training}.
Although we reproduce the simple adversarial training approaches here, a direct comparison of how these results compare to those reported by \citep{tramer2019adversarial} can be found in Appendix \ref{app:tramer}. Furthermore, while adversarial defenses are generally not intended to be robust to attacks outside of the perturbation model, we show some experiments exploring this aspect in Appendix \ref{app:outside}, namely the performance on the CIFAR10-C dataset (CIFAR10 with common corruptions) as well as exploring what happens when one defends against only two adversaries and evaluates on a third, unseen adversary.


\paragraph{Dataset variability} In addition to converging to suboptimal trade-offs between different adversaries as seen on MNIST, we find that the performance of simpler versions of adversarial training for multiple perturbations can also vary significantly based on the dataset. While the \maxt{} approach performed better than \avgt{} on MNIST, in the CIFAR10 setting we find that these roles are swapped: the \maxt{} approach converged to a suboptimal local minima which is 5.7\% less robust against the union of perturbation models than \avgt{}. Once again, this highlights the inconsistency of the simpler generalizations of adversarial training: depending on the problem setting, they may converge to suboptimal local optima which do not minimize the robust optimization objective from Equation \eqref{eq:robust_opt_multiple}. On the other hand, in both problem settings, we find \MSD{} consistently converges to a local optimum which is better at minimizing the worst-case loss in the union of the perturbation models, achieving 47.0\% robust accuracy, improving upon the best-performing simpler method of \avgt{} by 6.4\%.

\section{Conclusion}
In this paper, we showed that previous approaches aimed towards learning models which are adversarially robust to multiple perturbation models can be highly variable (across parameters and datasets), and difficult to tune, thereby converging to suboptimal local minima with trade-offs which do not defend against the union of multiple perturbation models. 
On the other hand, by incorporating the different perturbation models directly into the direction of steepest descent, our proposed approach of \MSD{} consistently outperforms past approaches across both MNIST and CIFAR10. The approach inherits the scalability and generality of adversarial training, without relying on specific complex architectures, and is able to better accomplish the robust optimization objective. We recommend using \MSD{} to directly minimize the worst-case performance among multiple perturbation models. 

\section*{Acknowledgements}
Eric Wong was funded by support from the Bosch Center for AI, under contract 0087016732PCR, and a fellowship from the Siebel Scholars Foundation.
Pratyush Maini was supported by a fellowship from the Khorana Program for Scholars, aided jointly by the Department of Science \& Technology Govt. of India and the US Department of State.

\bibliography{icml2020}
\bibliographystyle{icml2020}

\clearpage

\appendix

\section*{Adversarial Robustness Against the Union of Multiple Perturbation Models\\ (Supplementary Material)}

\section{Steepest descent and projections for $\ell_\infty$, $\ell_2$, and $\ell_1$ adversaries}
\label{app:steep}
In this section, we describe the steepest descent and projection steps for $\ell_p$ adversaries for 
$p \in \{\infty, 2, 1\}$; these are standard results, but are included for a complete description of the algorithms.  
Note that this differs slightly 
from the adversaries considered in \citet{schott2018towards}: while they used an $\ell_0$ adversary, 
we opted to use an $\ell_1$ adversary. The $\ell_0$ ball with radius $\epsilon$ is contained within an $\ell_1$ ball with the same radius, so achieving robustness against an $\ell_1$ adversary is strictly more difficult. 

\paragraph{$\ell_\infty$ space}
The direction of steepest descent with respect to the $\ell_\infty$ norm is 
\begin{equation}
    v_\infty(\delta) = \alpha \cdot \mathrm{sign}(\nabla l(x + \delta; \theta))
\end{equation}
and the projection operator onto $\Delta_{\infty, \epsilon}$ is  
\begin{equation}
    \mathcal P_{\Delta_{\infty, \epsilon}}(\delta) = \mathrm{clip}_{[-\epsilon, \epsilon]}(\delta)
\end{equation}

\paragraph{$\ell_2$ space}
The direction of steepest descent with respect to the $\ell_2$ norm is  
\begin{equation}
    v_2(\delta) = \alpha \cdot \frac{\nabla \ell(x + \delta; \theta)}{\|\nabla \ell(x + \delta; \theta)\|_2}
\end{equation}
and the projection operator onto the $\ell_2$ ball around $x$ is 
\begin{equation}
    \mathcal P_{\Delta_{2, \epsilon}}(\delta) =  \epsilon \cdot \frac{\delta}{\max\{\epsilon, \|\delta\|_2\}}
\end{equation}

\paragraph{$\ell_1$ space}
The direction of steepest descent with respect to the $\ell_1$ norm is 
\begin{equation}
    v_1(\delta) = \alpha \cdot \mathrm{sign}\left (\frac{\partial \ell(x + \delta; \theta)}{\partial \delta_{i^\star}} \right ) \cdot e_{i^\star}
\end{equation}
where
\begin{equation}
    i^\star =  \argmax_i |\nabla l(x + \delta; \theta)_i| 
    \label{eq:l1_coordinate}
\end{equation}
and $e_{i^*}$ is a unit vector with a one in position $i^*$. 
Finally, the projection operator onto the $\ell_1$ ball, 
\begin{equation}
    \mathcal P_{\Delta_{1, \epsilon}}(\delta) =  \argmin_{\delta' : \| \delta'\|_1 \leq \epsilon} \| \delta - \delta' \|_2^2, 
\end{equation}
can be solved with Algorithm \ref{alg:proj_l1}, and we refer the reader to  \citet{Duchi:2008:EPL:1390156.1390191} for its derivation.

\begin{algorithm}[t]
  \caption{Projection of some perturbation $\delta\in \mathbb R^n$ onto the $\ell_{1}$ ball with radius $\epsilon$. We use $|\cdot|$ to denote element-wise absolute value. }
    \begin{algorithmic}
  \STATE {\bfseries Input:} perturbation $\delta$, radius $\epsilon$
  \STATE Sort $|\delta|$ into $\gamma$ : $\gamma_{1}\geq\gamma_{2}\geq\dots\geq\gamma_{n}$
  \STATE $\rho \coloneqq \max \left\{ j \in \left[n \right ] : \gamma_{j} - \frac{1}{j}  \left(\sum_{r = 1}^{j}  \gamma_{r} - \epsilon\right)> 0  \right \}$
  \STATE $\eta \coloneqq \frac{1}{\rho}  \left(\sum_{i = 1}^{\rho}  \gamma_{i} - \epsilon\right)$
  \STATE $z_{i} \coloneqq \sign(\delta_i)\max\left\{ \gamma_{i}- \eta, 0 \right\}$ for $i = 1\dots n$
  \STATE {\bfseries return} $z$
\end{algorithmic}
\label{alg:proj_l1}
\end{algorithm}

\subsection{Enhanced $\ell_1$ steepest descent step} 
\label{app:pgd_l1_topk} 
Note that the steepest descent step for $\ell_1$ only updates a single coordinate per step. This can be quite inefficient, as pointed out by \citet{tramer2019adversarial}. To tackle this issue, and also empirically improve the attack success rate, \citet{tramer2019adversarial} instead select the top $k$ coordinates according to Equation \refeq{eq:l1_coordinate} to update. In this work, we adopt a similar but slightly modified scheme: we randomly sample $k$ to be some integer within some range $[k_1,k_2]$, and update each coordinate with step size $\alpha'= \alpha/k$. We observe in our experimentation that the randomness induced by varying the number of coordinates aids in reducing the gradient masking problem observed by \citet{tramer2019adversarial}. 

\subsection{Restricting the steepest descent coordinate} The steepest descent direction for both the $\ell_0$ and $\ell_1$ norm end up selecting a single coordinate direction to move the perturbation. However, if the perturbation is already at the boundary of pixel space (for MNIST, this is the range [0,1] for each pixel), then it's possible for the PGD adversary to get stuck in a loop trying to use the same descent direction to escape pixel space. To avoid this, we only allow the steepest descent directions for these two attacks to choose coordinates that keep the image in the range of real pixels.
\section{Extended results}
\label{app:evaluation}
Here, we show the full break down of adversarial error rates over individual attacks for both MNIST and CIFAR10.

\begin{table*}
  \caption{Summary of adversarial accuracy results for MNIST}
  \label{table:mnist_extended}
  \centering

  \begin{tabular}{l|rrrrrrrr}
    \hline
                                            &$P_\infty$& $P_2$ & $P_1$ & B-ABS & ABS & \maxt{} & \avgt{} & \MSD{} \\
                                             
    \hline
    Clean Accuracy                           & 99.1\% & 99.2\% & 99.3\% & 99\% & 99\% & 98.6\% & 99.1\% & 98.3\% \\
    \hline
    PGD-$\ell_\infty$                        & 90.3\% & 0.4\%  & 0.0\% &    -  &    - & 51.0\% & 65.2\% & 62.7\% \\
    FGSM                                     & 94.9\% & 68.3\% & 6.4\% & 85\%  & 34\% & 81.4\% & 85.5\% & 82.8\% \\
    PGD-Foolbox                              & 92.1\% & 8.5\%  & 0.1\% & 86\%  & 13\% & 65.8\% & 73.5\% & 69.2\% \\
    MIM                                      & 92.3\% & 11.2\% & 0.1\% & 85\%  & 17\% & 70.7\% & 76.7\% & 71.0\% \\
    \hline
    $\ell_\infty$ attacks $(\epsilon=0.3)$   & 90.3\% & 0.4\% & 0.0\% & 77\%   &  8\% & 51.0\% & 65.2\% & 62.7\% \\
    \hline
    PGD-$\ell_2$                             & 68.8\% & 69.2\% & 38.7\% &  -   &  -   & 64.1\% & 67.9\% & 70.2\% \\
    PGD-Foolbox                              & 88.9\% & 77.9\% & 48.7\% & 63\% & 87\% & 75.6\% & 80.3\% & 78.4\% \\
    Gaussian Noise                           & 98.9\% & 98.6\% & 98.9\% & 89\% & 98\% & 97.7\% & 98.6\% & 97.2\% \\
    Boundary Attack                          & 18.2\% & 81.4\% & 62.1\% & 91\% & 83\% & 73.6\% & 71.8\% & 72.4\% \\
    DeepFool                                 & 93.0\% & 86.8\% & 59.5\% & 41\% & 83\% & 81.7\% & 87.3\% & 80.7\% \\
    Pointwise Attack                         & 40.6\% & 95.1\% & 96.7\% & 87\% & 94\% & 90.8\% & 85.9\% & 89.6\% \\
    DDN                                      & 63.9\% & 70.5\% & 40.0\% &  -   &  -   & 62.5\% & 64.6\% & 69.5\% \\
    CWL2                                     & 79.6\% & 74.5\% & 44.8\% &  -   &  -   & 72.1\% & 72.4\% & 74.5\% \\    
    \hline
    $\ell_2$ attacks $(\epsilon=2.0)$        & 13.6\% & 69.2\% & 38.5\% &39\%  &80\%  & 61.9\% & 60.1\% & 67.9\% \\
    \hline
    PGD-$\ell_1$                             & 61.8\% & 51.1\% & 74.6\% &  -   &  -   & 61.2\% & 66.5\% & 70.4\% \\
    Salt \& Pepper                           & 62.1\% & 96.4\% & 97.7\% & 96\% & 95\% & 94.6\% & 90.6\% & 89.1\% \\
    Pointwise Attack                         & 5.3\%  & 83.3\% & 89.1\% & 82\% & 78\% & 65.3\% & 45.4\% & 70.7\% \\
    \hline
    $\ell_1$ attacks $(\epsilon=10)$         & 4.2\%  & 43.4\% & 70.0\% &82\%  & 78\% & 52.6\% & 39.2\% & 65.0\% \\
    \hline
    All attacks                              & 3.7\%  & 0.4\%  & 0.0\%  & 39\% &  8\% & 42.1\% & 34.9\% & \textbf{58.4}\% \\
    \hline
  \end{tabular}
\end{table*}

\subsection{MNIST results}
\label{app:mnist}
\paragraph{Expanded table of results}
Table \ref{table:mnist_extended} contains brak down of adversarial accuracies against all attacks for all models on the MNIST dataset. All attacks were run on a subset of the first 1000 test examples with 10 random restarts, with the exception of Boundary Attack, which by default makes 25 trials per iteration, and DDN attack, which does not benefit from restarts owing to a deterministic starting point. 
The results for B-ABS and ABS models are reported directly from \citet{schott2018towards}, which uses gradient estimation techniques whenever a gradient is needed, and the robustness against all attacks for B-ABS and ABS is an upper bound based on the reported results. Further, they used epsilon balls of radii (0.3,1.5,12) for ($\ell_\infty$, $\ell_2$, $\ell_0$) adversaries. Moreover, they used an $\ell_0$ perturbation model of a higher radius and evaluated against $\ell_0$ attacks. So the reported number is a near estimate of the $\ell_1$ adversarial accuracy.

\subsection{CIFAR10 results}
\label{app:cifar10}
\paragraph{Expanded table of results}
Table \ref{table:cifar10_extended} contains the full table of results for all attacks on all models on the CIFAR10 dataset. All attacks were run on a subset of the first 1000 test examples with 10 random restarts, with the exception of Boundary Attack, which by default makes 25 trials per iteration, and DDN attack, which does not benefit from restarts owing to a deterministic starting point. Further note that salt \& pepper and pointwise attacks in the $\ell_1$ section are technically $\ell_0$ attacks, but produce perturbations in the $\ell_1$ ball. Finally, it is clear here that while the training against an $\ell_1$ PGD adversary defends against said PGD adversary, it does not seem to transfer to robustness against other attacks. 
\begin{table*}
  \caption{Summary of adversarial accuracy results for CIFAR10}
  \label{table:cifar10_extended}
  \centering
  \begin{tabular}{l|rrrrrr}
    \hline
                                         & $P_\infty$ & $P_2$ & $P_1$ & \maxt{} & \avgt{} & \MSD{} \\
    \hline
    Clean accuracy                          & 83.3\% & 90.2\% & 73.3\% & 81.0\% & 84.6\% & 81.1\% \\
    \hline
    PGD-$\ell_\infty$                        & 50.3\% & 48.4\% & 29.8\% & 44.9\% & 42.8\% & 48.0\%\\
    FGSM                                     & 57.4\% & 43.4\% & 12.7\% & 54.9\% & 51.9\% & 53.7\%\\
    PGD-Foolbox                              & 52.3\% & 28.5\% & 0.6\% & 48.9\% & 44.6\% & 53.5\%\\
    MIM                                      & 52.7\% & 30.4\% & 0.7\% & 49.9\% & 46.1\% & 50.7\%\\
    \hline
    $\ell_\infty$ attacks $(\epsilon=0.03)$ & 50.7\% & 28.3\% & 0.2\% & 44.9\% & 42.5\% & 48.0\%\\
    \hline
    PGD-$\ell_2$                             & 59.0\% & 62.1\% & 28.9\% & 64.1\% & 66.9\% & 66.6\%\\
    PGD-Foolbox                              & 61.6\% & 64.1\% & 4.9\% & 65.0\% & 68.0\% & 68.2\%\\
    Gaussian Noise                           & 82.2\% & 89.8\% & 62.3\% & 81.3\% & 84.3\% & 80.9\%\\
    Boundary Attack                          & 65.5\% & 67.9\% & 2.3\% & 64.4\% & 69.2\% & 69.4\%\\
    DeepFool                                 & 62.2\% & 67.3\% & 0.9\% & 64.4\% & 67.4\% & 66.1\%\\
    Pointwise Attack                         & 80.4\% & 88.6\% & 46.2\% & 78.9\% & 83.8\% & 79.8\%\\
    DDN                                      & 60.0\% & 63.5\% & 0.1\% & 64.5\% & 67.7\% & 67.0\%\\
    CWL2                                     & 62.0\% & 71.6\% & 0.1\% & 66.9\% & 71.5\% & 64.7\%\\
    \hline
    $\ell_2$ attacks $(\epsilon=0.05)$       & 57.3\% & 61.6\% & 0.0\% & 61.7\% & 65.0\% & 64.3\%\\
    \hline
    PGD-$\ell_1$                             & 16.5\% & 49.2\% & 69.1\% & 39.5\% & 54.0\% & 53.4\%\\
    Salt \& Pepper                           & 63.4\% & 74.2\% & 35.5\% & 75.2\% & 80.7\% & 73.9\%\\
    Pointwise Attack                         & 49.6\% & 62.4\% & 8.4\% & 63.3\% & 77.0\% & 69.7\%\\
    \hline
    $\ell_1$ attacks $(\epsilon=12)$         & 16.0\% & 46.6\% & 7.9\% & 39.4\% & 54.0\% & 53.0\%\\
    \hline
    All attacks                              & 15.6\% & 27.5\% & 0.0\% & 34.9\% & 40.6\% & \textbf{47.0}\%\\
    \hline
  \end{tabular}
\end{table*}

\section{Experimental details}
\label{app:setup}

\subsection{Hyperparameters for PGD adversaries}
\label{app:hp_adversary}
In this section, we describe the parameters used for all PGD adversaries in this paper. 
\paragraph{MNIST} 

The $\ell_\infty$ adversary used a step size $\alpha=0.01$ within a radius of $\epsilon=0.3$ for 50 iterations. 

The $\ell_2$ adversary used a step size $\alpha=0.1$ within a radius of $\epsilon=2.0$ for 100 iterations. 

The $\ell_1$ adversary used a step size of $\alpha=0.8$ within a radius of $\epsilon=10$ for 50 iterations. By default the attack is run with two restarts, once starting with $\delta$ = 0 and once by randomly initializing $\delta$ in the allowable perturbation ball. $k_1$ = 5, $k_2$ = 20 as described in \ref{app:pgd_l1_topk}.

At test time, we increase the number of iterations to $(100,200,100)$ for $(\ell_\infty, \ell_2, \ell_1)$. 

\paragraph{CIFAR10} \;

The $\ell_\infty$ adversary used a step size $\alpha=0.003$ within a radius of $\epsilon=0.03$ for 40 iterations. 

The $\ell_2$ adversary used a step size $\alpha=0.05$ within a radius of $\epsilon=0.5$ for 50 iterations. 

The $\ell_1$ adversary used a step size  $\alpha=1.0$ with $\epsilon=12$ for 50 iterations. $k_1$ = 5, $k_2$ = 20 as described in \ref{app:pgd_l1_topk}.

At test time, we increase the number of iterations to $(100,500,100)$ for $(\ell_\infty, \ell_2, \ell_1)$. 

\subsection{Training hyperparameters}
\label{app:hp_adv_training}
In this section, we describe the parameters used for adversarial training. 

\paragraph{MNIST} For all the models, we used the Adam optimizer without weight decay, and used a variation of the learning rate schedule from \citet{smith2018disciplined}, which is piecewise linear from 0 to $10^{-3}$ over the first 6 epochs, and down to 0 over the last 9 epochs.

We perform a large hyperparameter search for each of the  \maxt{}, \avgt{}, \MSD{} models, by training them for 15 epochs on all combinations of the following step sizes: 
$\alpha_1$ = \{0.75, 0.8, 1.0, 2.0\}, $\alpha_2$ = \{0.1, 0.2\}, $\alpha_\infty$ = \{0.01, 0.02, 0.03\}. Also, we find that setting the maximum value of learning rate to $10^{-3}$ works best among other values that we experiment on.

The \MSD{} adversary used step sizes of $\alpha=(0.01, 0.1, 0.8)$ for the $(\ell_\infty, \ell_2, \ell_1)$ directions within a radius of $\epsilon=(0.3,2.0,10)$ for 100 iterations.

The \maxt{} approach used step sizes of $\alpha=(0.01, 0.1, 1.0)$ for the $(\ell_\infty, \ell_2, \ell_1)$ directions within a radius of $\epsilon=(0.3,2.0,12)$ for (50, 100, 100) iterations respectively. We had to make an early stop at the end of the fourth epoch, since further training made the model biased towards $\ell_\infty$ robustness. We also had to increase the number of restarts and attack iterations for the $\ell_1$ PGD attack.

The \avgt{} approach used step sizes of $\alpha=(0.01, 0.2, 1.0)$ for the $(\ell_\infty, \ell_2, \ell_1)$ directions within a radius of $\epsilon=(0.3,2.0,12)$ for (50, 100, 50) iterations respectively. Note that we had to change the perturbation model for the $\ell_1$ adversary to make it relatively stronger in-order to ``balance" the trade-offs between different perturbation models.

Finally, we train the standard $P_1$, $P_2$, $P_\infty$ models for an extended period till 20 epochs with respective step sizes $\alpha_1$ = 1.0, $\alpha_2$ = 0.1, and $\alpha_\infty$ = 0.01.

\begin{table*}[t]
  \caption{Comparison with \citet{tramer2019adversarial} on MNIST (higher is better). Results for all models except \MSD{} are taken as is from \citet{tramer2019adversarial}}
  \label{table:tramer:mnist}
  \centering
  \begin{tabular}{l|rrrrrrr}
    \hline
                                    &Vanilla & $Adv_{\infty}$ &$Adv_{1}$ &$Adv_{2}$& $Adv_{\avgt{}}$ & $Adv_{\maxt{}}$ & \textbf{\MSD{}} \\
    \hline
    Clean accuracy                           & 99.4\% & 99.1\% & 98.9\%  & 98.5\% & 97.3\% & 97.2\% & 98.3\% \\
    \hline
    $\ell_\infty$ attacks $(\epsilon=0.3)$   & 0.0\% & 91.1\% & 0.0\%   & 0.4\%  & 76.7\% & 71.7\% & 75.9\%\\
    $\ell_2$ attacks $(\epsilon=2.0)$        & 12.4\% & 12.1\% & 50.6\% & 71.8\% & 58.3\% & 56.0\% & 67.9\%\\
    $\ell_1$ attacks $(\epsilon=10)$         & 8.5\% & 11.3\% & 78.5\%  & 68.0\% & 53.9\% & 62.6\% & 74.8\%\\
    \hline
    All attacks                              & 0.0\% & 6.8\% & 0.0\% &0.4\% & 49.9\% & 52.4\% &\textbf{65.2}\%\\
    \hline
  \end{tabular}
\end{table*}
\begin{table*}[t]
  \caption{Comparison with \citet{tramer2019adversarial} on CIFAR10 (higher is better). Results for all models except \MSD{} are taken as is from \cite{tramer2019adversarial}}
  \label{table:tramer:cifar10}
  \centering
  \begin{tabular}{l|rrrrrr}
    \hline
                                    &Vanilla & $Adv_{\infty}$ &$Adv_{1}$ &$Adv_{\avgt{}}$ & $Adv_{\maxt{}}$ & \textbf{\MSD{}} \\
    \hline
    Clean accuracy                                      & 95.7\% & 92.0\% & 90.8\% & 91.1\% & 91.2\% & 92.0\% \\
    \hline
    $\ell_\infty$ attacks $(\epsilon=\frac{4}{255})$   & 0.0\% & 71.0\% & 53.4\%  & 64.1\% & 65.7\% & 66.8\%\\
    $\ell_1$ attacks $(\epsilon=\frac{2000}{255})$     & 0.0\% & 16.4\% & 66.2\%  & 60.8\% & 62.5\% & 65.3\%\\
    \hline
    All attacks                                         & 0.0\% & 16.4\% & 53.1\%  & 59.4\% & 61.1\% &\textbf{63.2}\%\\
    \hline
  \end{tabular}
\end{table*}

\paragraph{CIFAR10} For all the models, we used the SGD optimizer with momentum 0.9 and weight decay $5\cdot 10^{-4}$. We used a variation of the learning rate schedule from \citet{smith2018disciplined} to achieve superconvergence in 50 epochs, which is piecewise linear from 0 to 0.1 over the first 20 epochs, down to 0.005 over the next 20 epochs, and finally back down to 0 in the last 10 epochs. 

The \MSD{} adversary used step sizes of $\alpha=$ (0.003, 0.02, 1.0) for the $(\ell_\infty, \ell_2, \ell_1)$ directions within a radius of $\epsilon=(0.03,0.5,12)$ for 50 iterations. 

The \maxt{} adversary used step sizes of $\alpha=$ (0.005, 0.05, 1.0) for the $(\ell_\infty, \ell_2, \ell_1)$ directions within a radius of $\epsilon=(0.03,0.3,12)$ for (40, 50, 50) iterations respectively. We do an early stop at epoch 45 for best accuracy.

The \avgt{} adversary used step sizes of $\alpha=(0.003, 0.05, 1.0)$ for the $(\ell_\infty, \ell_2, \ell_1)$ directions within a radius of $\epsilon=(0.03,0.3,12)$ for (40, 50, 50) iterations respectively. 

\textbf{Note:} For obtaining the best-performing \maxt{} and \avgt{} models, we artificially balance the size of the $\ell_2$ perturbation region, reducing its radius to 0.3 from the actual threat model of radius 0.5.

\section{Comparison with \citet{tramer2019adversarial}}
\label{app:tramer}

In this section, we compare the results of our trained \MSD{} model with that of \citet{tramer2019adversarial}, who study the theoretical and empirical trade-offs of adversarial robustness in various settings when defending against multiple adversaries. Training methods presented by them in their comparisons, namely $Adv_{\avgt{}}$ and $Adv_{\maxt{}}$ closely resemble the simpler approaches discussed in this paper: \avgt{} and \maxt{} respectively. We use the results as is from their work, and additionally compare the position of our \MSD{} models at the revised thresholds used by \citet{tramer2019adversarial}. We make our best attempt at replicating the same attack strengths as of those used in the evaluation in \citet{tramer2019adversarial}. We use all attacks from the Foolbox library, apart from the PGD $\ell_1$ or SLIDE attack \citep{tramer2019adversarial}. Further, we do not make multiple random restarts for these comparisons, which is in line with their evaluation.

The results of Tables \ref{table:tramer:mnist} and \ref{table:tramer:cifar10} show that the relative advantage of \MSD{} over simpler techniques does hold up. The \MSD{} model was not retrained for the comparison on the MNIST dataset since it was trained to be robust to the same perturbation region in the main paper as well.

In case of CIFAR10, we train a model using the WideResNet architecture \cite{zagoruyko2016wide} with 5 residual blocks and a widening factor of 10, as used by \citet{tramer2019adversarial}. It may be noted that this model has 4 times more parameters than the pre-activation version of ResNet which was used for the comparisons in the main paper. Further, for the CIFAR10 results in Table \ref{table:tramer:cifar10}, the models are trained and tested only for $\ell_\infty$ and $\ell_1$ adversarial perturbations with $\epsilon$ = ($\frac{4}{255}$, $\frac{2000}{255}$) $\sim$(0.0157, 7.84). Note that the size of the perturbation regions considered in the main paper is strictly larger than these perturbation regions.

We emphasize that the evaluation method adopted in the main paper is stronger than that in this comparison. This may also be noted from the results in Table~\ref{table:tramer:mnist}, where the same \MSD{} model (without retraining) achieves nearly 7\% higher accuracy of 65.2\% against all attacks that were considered by \citet{tramer2019adversarial}, while the same model achieved an overall robust accuracy of 58.4\% in our evaluation in Table~\ref{table:mnist} in the main paper. These differences can be largely attributed to:
\begin{enumerate}
\item \textbf{Use of random restarts:} We observe in our experiments that using up to 10 restarts for all our attacks leads to a decrease in model accuracy from 5 to 10\% across all models. \citeauthor{tramer2019adversarial} do not mention restarting their attacks for these models and so the robust accuracies for their models in Tables \ref{table:tramer:mnist}, \ref{table:tramer:cifar10} could potentially be lowered with random restarts. 
\item \textbf{Larger Suite of Attacks Used:} The attacks used by \citeauthor{tramer2019adversarial} in case of the CIFAR10 dataset are PGD, EAD \citep{chen2017ead} and Pointwise Attack \citep{schott2018towards} for $\ell_1$; PGD, C\&W \citep{carlini2017towards} and Boundary Attack \citep{brendel2017decision} for $\ell_2$; and PGD for $\ell_\infty$. We use a more expansive suite of attacks as shown in Appendix \ref{app:evaluation}. Some attacks like DDN, which proved to be strong adversaries in most cases, were not considered by them.
\end{enumerate}
Our observations re-emphasize the importance of performing multiple restarts and using a broad suite of attacks in order to be able to best determine the robust performance of a proposed algorithm.

\section{Analyzing learned Filters for MNIST}
\label{app:filters}
\begin{figure*}[t]
  \centering
  \subcaptionbox{$P_\infty$ Model\label{fig:app:linf_filters}}{  \includegraphics[width=0.6\columnwidth]{figures/filters/linf_filters.pdf}}
  \hspace{0.2in}
  \subcaptionbox{$P_2$ Model\label{fig:l2_filters}}{  \includegraphics[width=0.6\columnwidth]{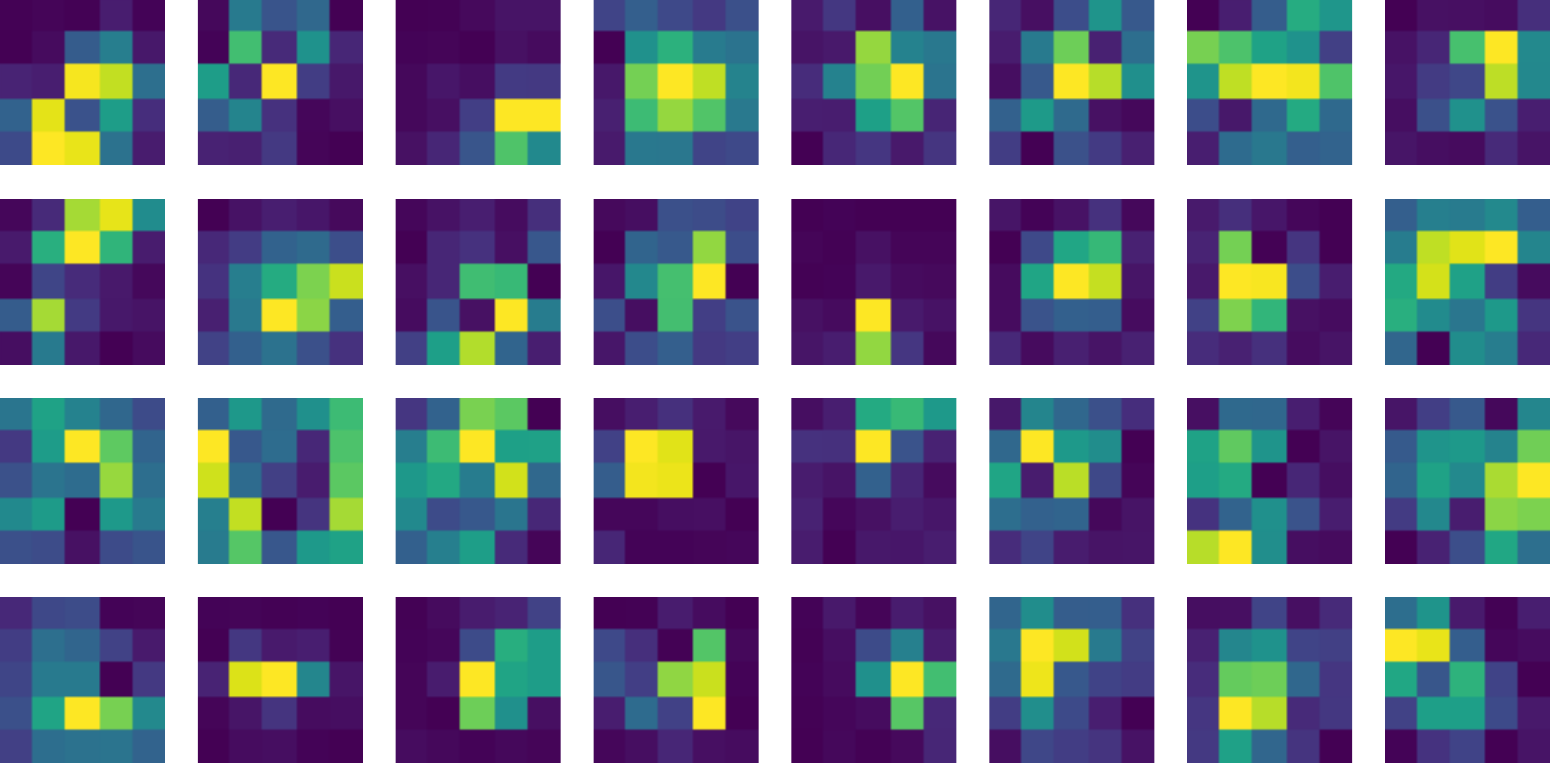}}
  \hspace{0.2in}
  \subcaptionbox{$P_1$ Model\label{fig:l1_filters}}{  \includegraphics[width=0.6\columnwidth]{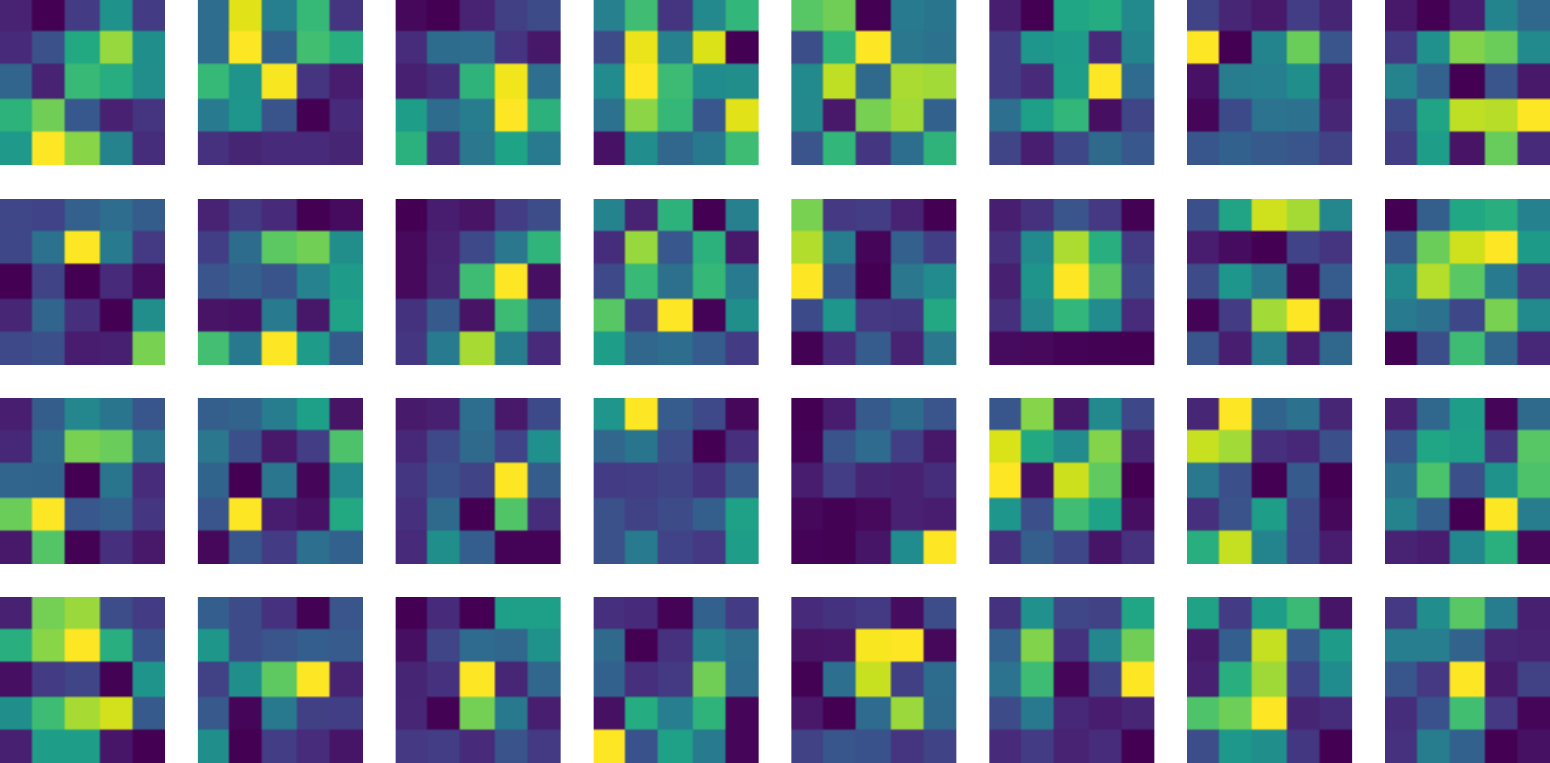}}
  \\
    \includegraphics[width=\columnwidth]{figures/filters/colorbar_1.pdf}
  \caption{A view of each of the (5x5) learned filters of the first layer of $P_\infty$, $P_2$, $P_1$ models trained on the MNIST dataset. While there are many learned filters in the $P_\infty$ model that have only one non-zero element (rest of the values are nearly zero), such a phenomenon is absent in $P_2$, $P_1$ models.}
  \label{app:fig:app_filters_1}
\end{figure*}

As described in \S~\ref{sec:Arch}, we use a simple 4 layer CNN model to classify MNIST digits. Each of the two convolutional layers has 5x5 filters. Specifically, the first layer contains 32 such filters.
We begin our analysis by observing the learned filters of an $\ell_\infty$ robust model. We observe that many of the learned filters are extremely sparse with only one non-zero element as shown in Figure \ref{fig:app:linf_filters}. Interestingly, such a view is unique to the case of the $\ell_\infty$ robust model and is not observed in $\ell_2$ (Figure \ref{fig:l2_filters}) and $\ell_1$ (Figure \ref{fig:l1_filters}) robust models.

\begin{figure*}[t]
  \centering
  \subcaptionbox{\MSD{} Model ($\alpha_1$ = 0.8, $\alpha_2$ = 0.1, $\alpha_\infty$ = 0.01)\label{fig:msd_1_filters}}{  \includegraphics[width=0.9\columnwidth]{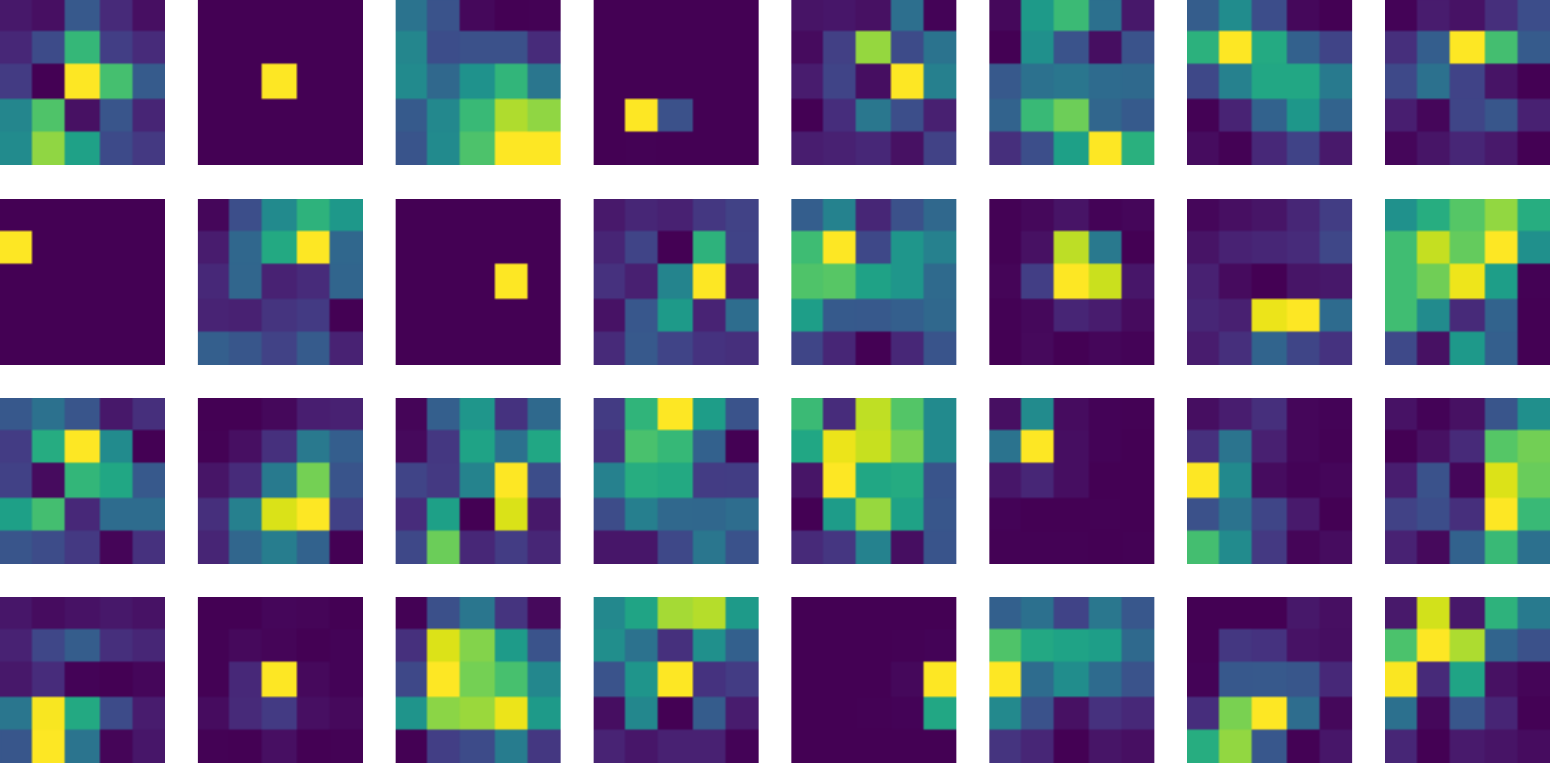}}
  \hspace{0.4in}
  \subcaptionbox{\MSD{} Model ($\alpha_1$ = 0.8, $\alpha_2$ = 0.1, $\alpha_\infty$ = 0.03)\label{fig:msd_2_filters}}{  \includegraphics[width=0.9\columnwidth]{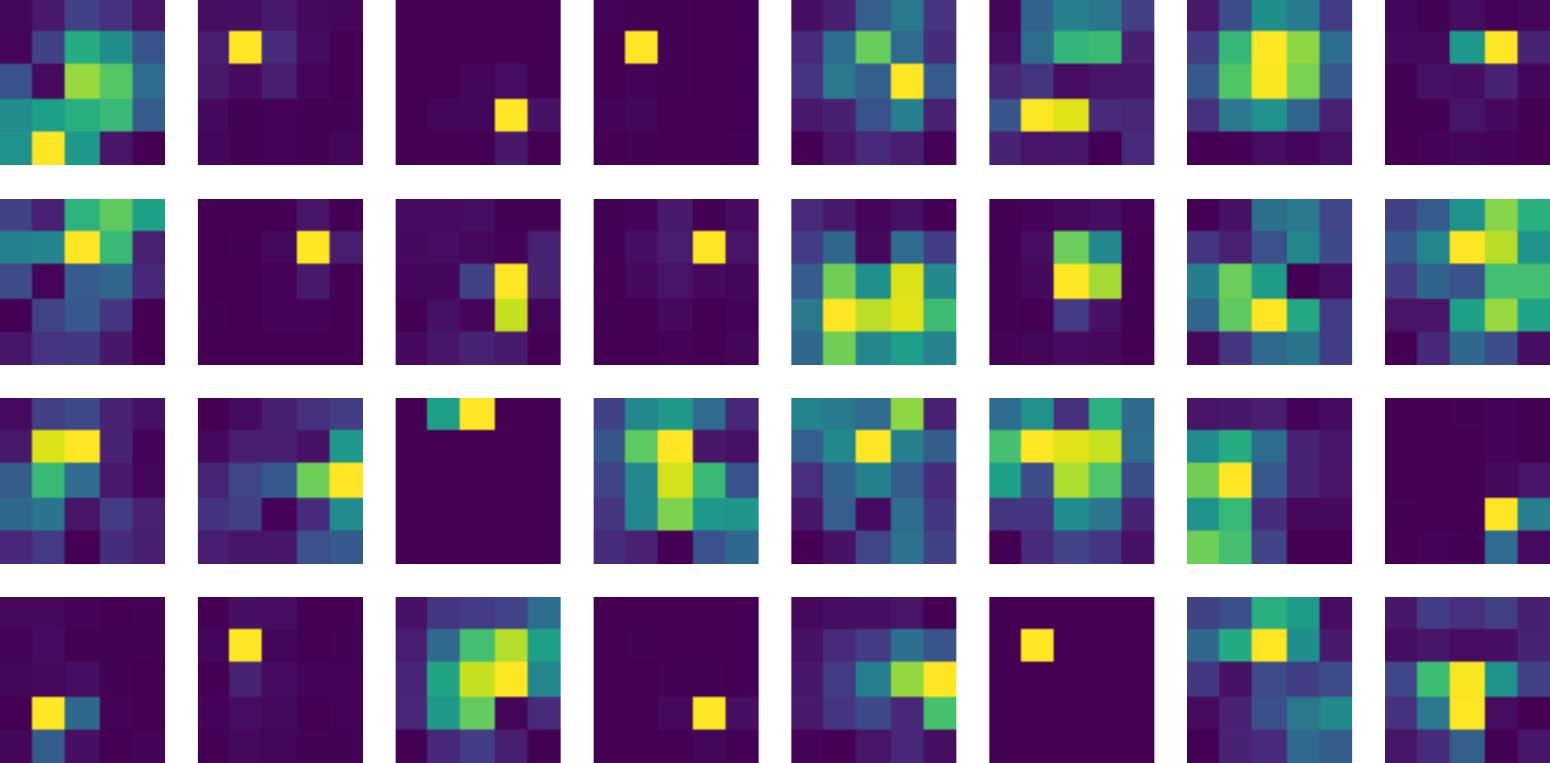}}
  \\
      \includegraphics[width=\columnwidth]{figures/filters/colorbar_1.pdf}
  \caption{A view of each of the (5x5) learned filters of the first layer of \MSD{} models trained on the MNIST dataset. The training hyper-parameters for the left and right images only differ in the step-size for the $\ell_\infty$ attack, where $\alpha_\infty$ = 0.01 for the left and $\alpha_\infty$ = 0.03 for the right image. The figure suggests how adjusting the relative step-sizes can help reduce the occurrence of sparse filters in case of \MSD{} models.}
  \label{app:fig:app_filters_2}
\end{figure*}

\begin{figure*}[t]
  \centering
  \subcaptionbox{\maxt{} Model ($\alpha_1$ = 0.8, $\alpha_2$ = 0.1, $\alpha_\infty$ = 0.01)\label{fig:max_1_filters}}{  \includegraphics[width=0.9\columnwidth]{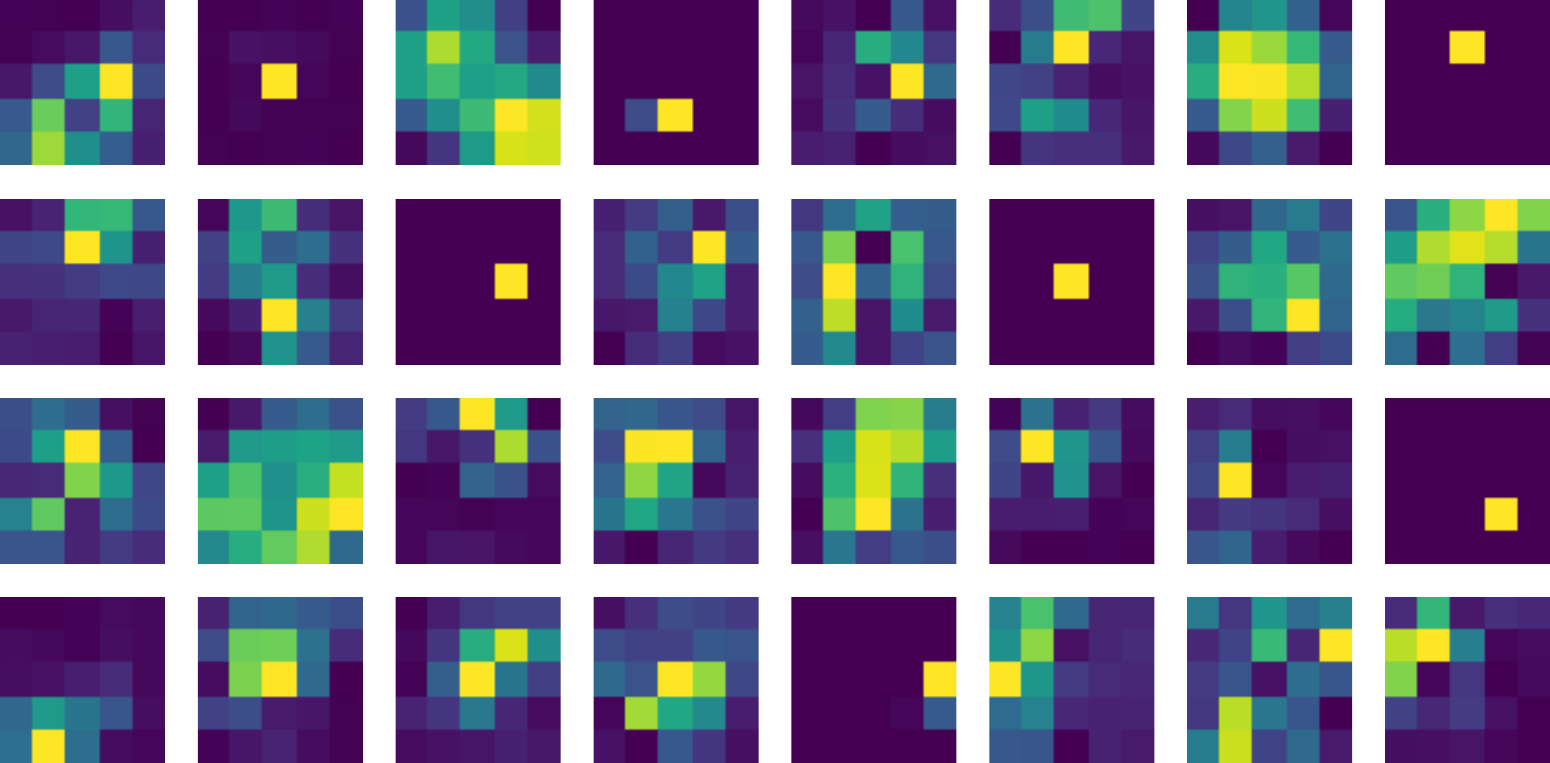}}
  \hspace{0.4in}
  \subcaptionbox{\maxt{} Model ($\alpha_1$ = 0.8, $\alpha_2$ = 0.1, $\alpha_\infty$ = 0.03)\label{fig:max_2_filters}}{  \includegraphics[width=0.9\columnwidth]{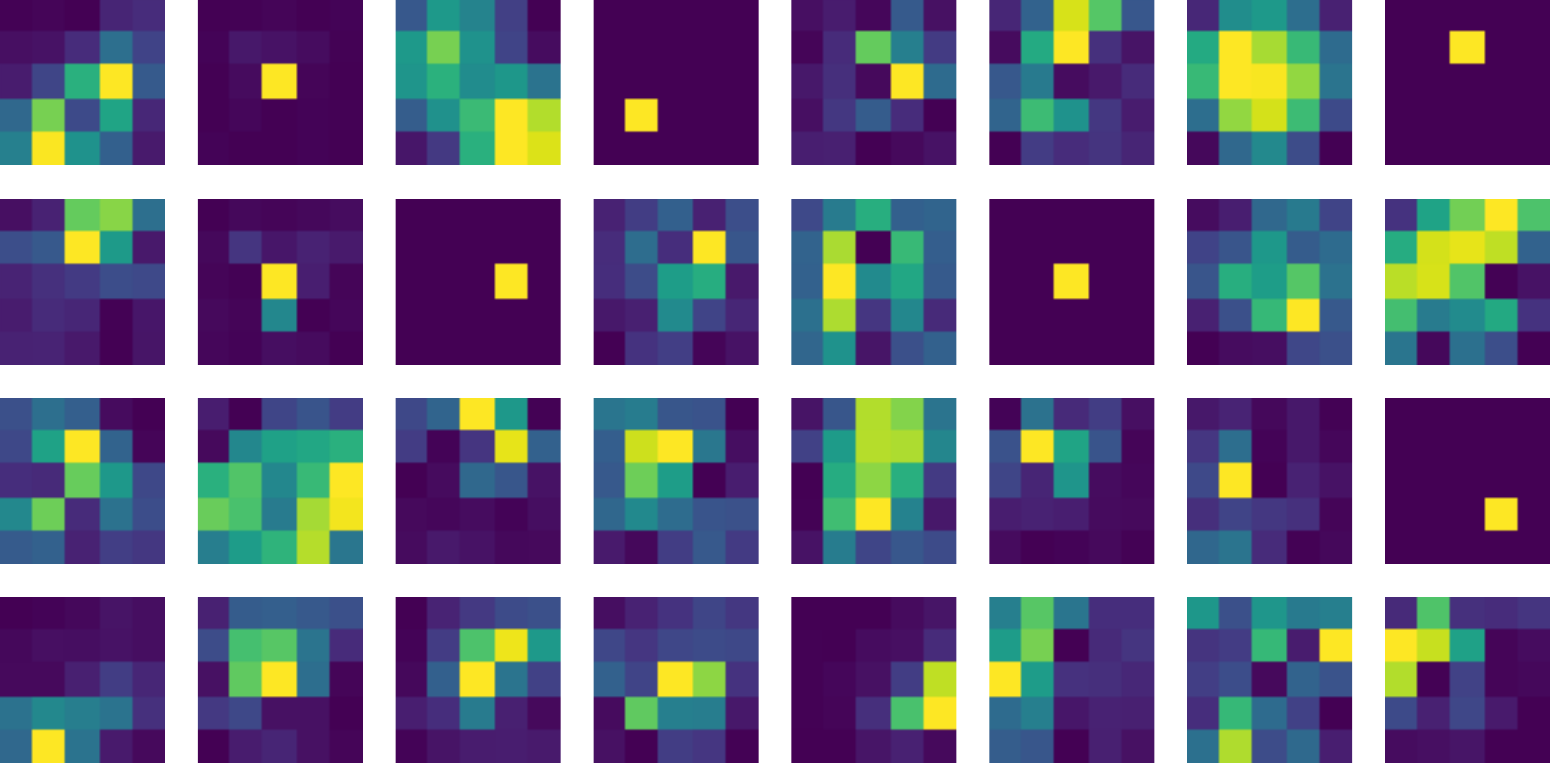}}
  \\
      \includegraphics[width=\columnwidth]{figures/filters/colorbar_1.pdf}
  \caption{A view of each of the (5x5) learned filters of the first layer of \maxt{} models trained on the MNIST dataset. The training hyper-parameters for the left and right images only differ in the step-size for the $\ell_\infty$ attack, where $\alpha_\infty$ = 0.01 for the left and $\alpha_\infty$ = 0.03 for the right image. The learned filters are nearly identical for both models and indicate how there may not be a natural way of balancing the trade-offs between different perturbation models in the training schedule for \maxt{} models.}
  \label{app:fig:app_filters_3}
\end{figure*}

\begin{figure*}[t]
  \centering
  \subcaptionbox{\avgt{} Model ($\alpha_1$ = 0.8, $\alpha_2$ = 0.1, $\alpha_\infty$ = 0.01)\label{fig:avg_1_filters}}{  \includegraphics[width=0.9\columnwidth]{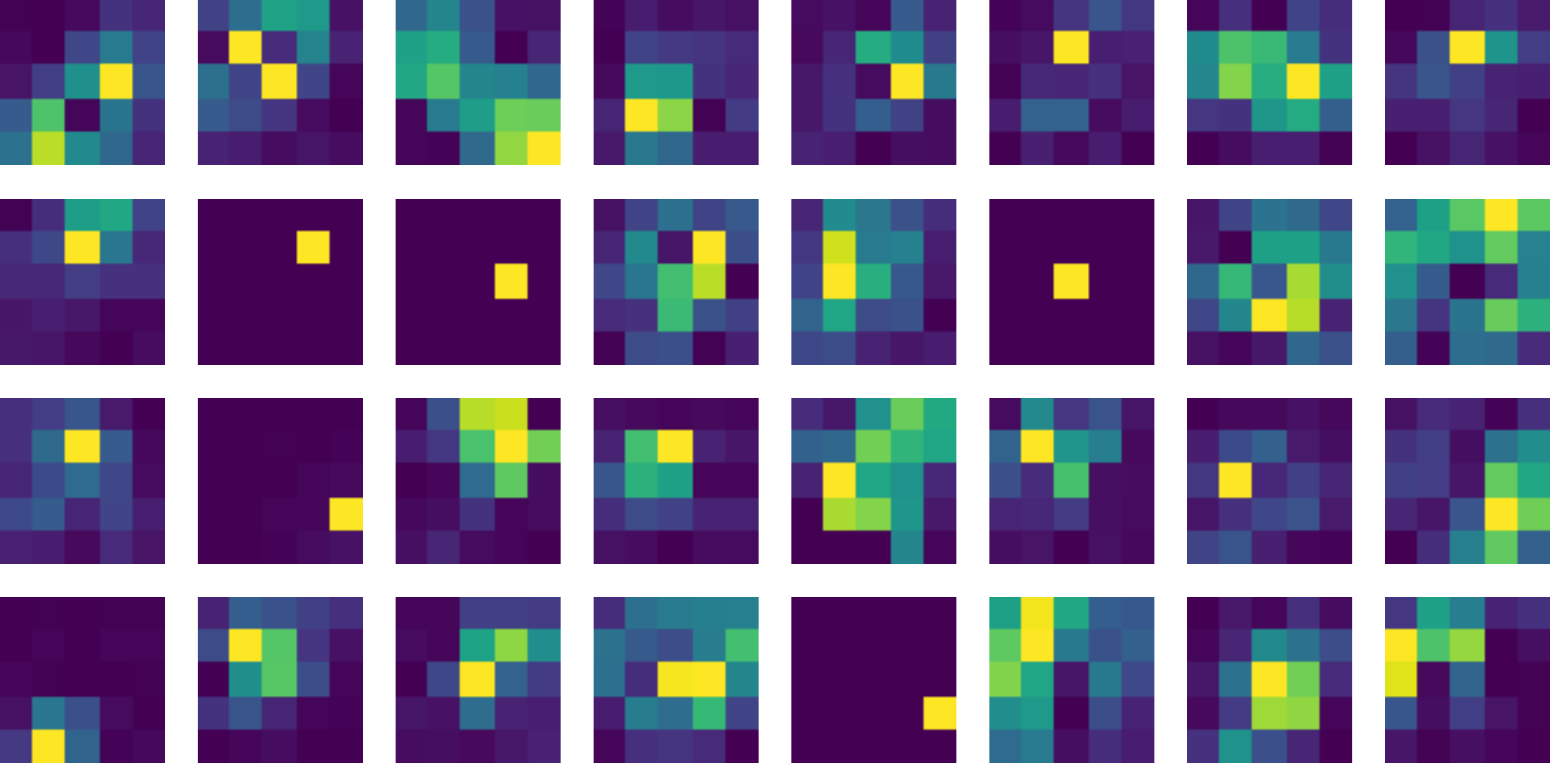}}
  \hspace{0.4in}
  \subcaptionbox{\avgt{} Model ($\alpha_1$ = 0.8, $\alpha_2$ = 0.1, $\alpha_\infty$ = 0.03)\label{fig:avg_2_filters}}{  \includegraphics[width=0.9\columnwidth]{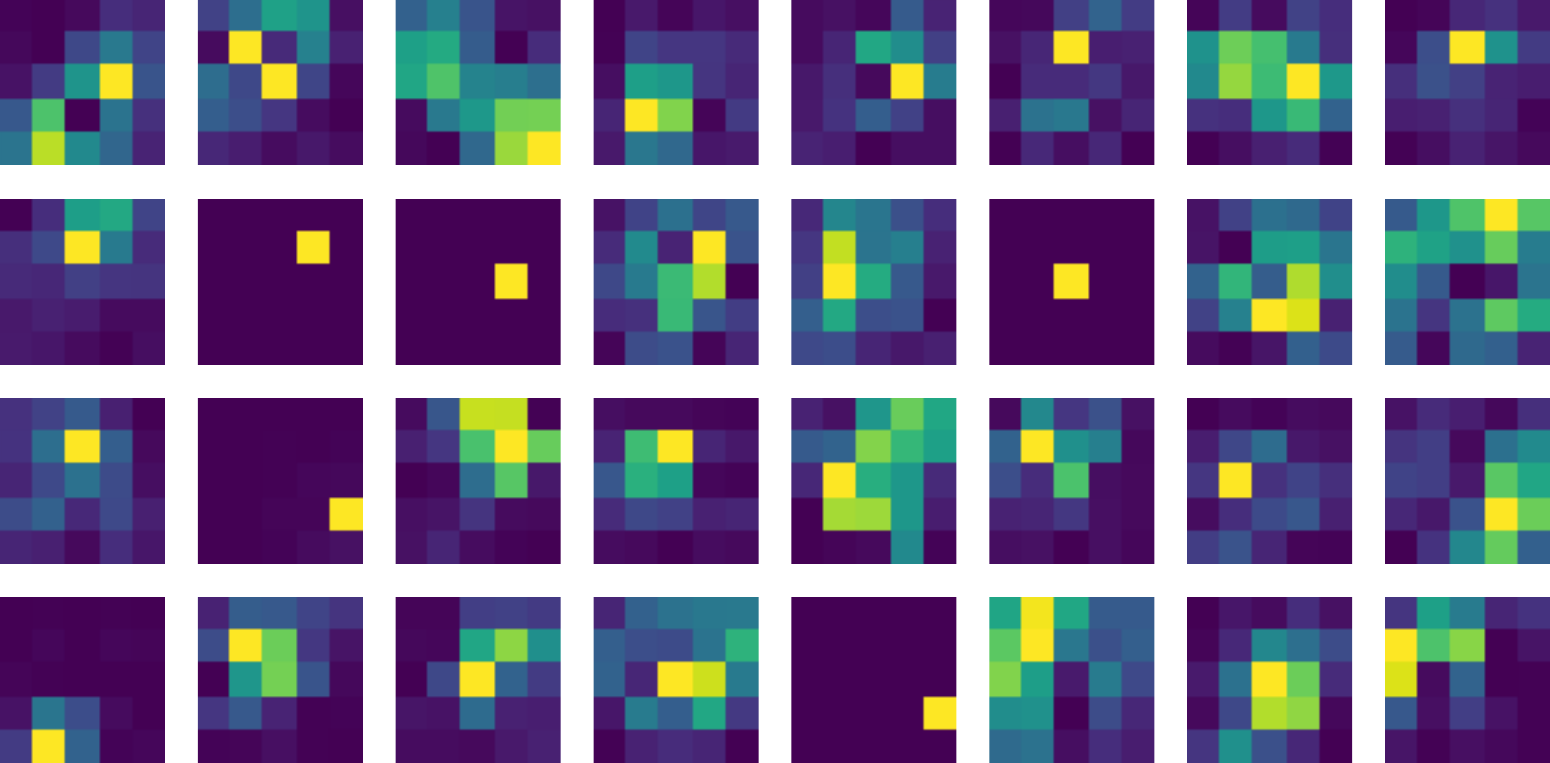}}
  \\
    \includegraphics[width=\columnwidth]{figures/filters/colorbar_1.pdf}
  \caption{A view of each of the (5x5) learned filters of the first layer of \avgt{} models trained on the MNIST dataset. The training hyper-parameters for the left and right images only differ in the step-size for the $\ell_\infty$ attack, where $\alpha_\infty$ = 0.01 for the left and $\alpha_\infty$ = 0.03 for the right image. The learned filters are nearly identical for both models and indicate how there may not be a natural way of balancing the trade-offs between different perturbation models in the training schedule for \avgt{} models.}
  \label{app:fig:app_filters_4}
\end{figure*}

The presence of such learned filters that act as thresholding filters, due to the immediately followed activation layer, has been hypothesized to be the reason for gradient masking in such models by \citet{madry2018towards,tramer2019adversarial}. The hypothesis is in line with our experimental correlations of $\ell_\infty$ model being the only standard model that performs poorly against decision-based adversaries while being significantly robust to first-order adversaries. Therefore, we go beyond this preliminary analysis to observe the initial layers of \MSD{} (Figures \ref{fig:msd_1_filters}, \ref{fig:msd_2_filters}), \maxt{} (Figures \ref{fig:max_1_filters}, \ref{fig:max_2_filters}), \avgt{} (Figures \ref{fig:avg_1_filters}, \ref{fig:avg_2_filters}) models. In all the three cases, we have two models that are almost identically trained, but with different $\ell_\infty$ step sizes: $\alpha_\infty$ = 0.01 on the left and $\alpha_\infty$ = 0.03 on the right.
While we display results only on two extreme settings of relative attack step-sizes, we find that changing the relative step size of different PGD adversaries can help reduce the number of thresholding filters in the \MSD{} approach, which also leads to better accuracies against decision-based attacks like the Pointwise Attack. However, the \maxt{} and \avgt{} models are nearly invariant to the individual attack step-sizes. 

\begin{figure*}[t]
  \centering
  \subcaptionbox{Final \maxt{} Model\label{fig:max_final_filters}}{  \includegraphics[width=0.9\columnwidth]{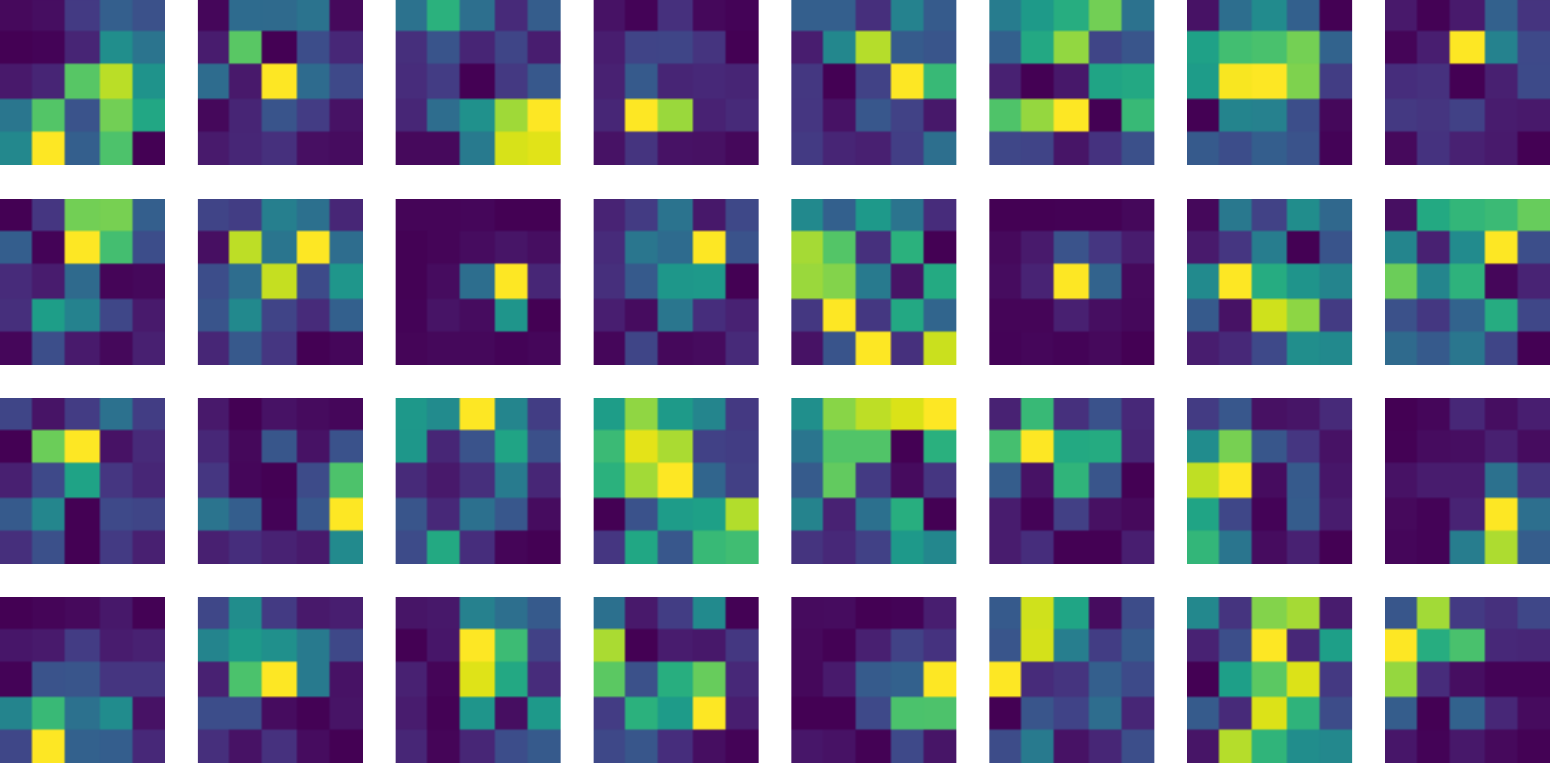}}
  \hspace{0.4in}
  \subcaptionbox{Final \avgt{} Model\label{fig:avg_final_filters}}{  \includegraphics[width=0.9\columnwidth]{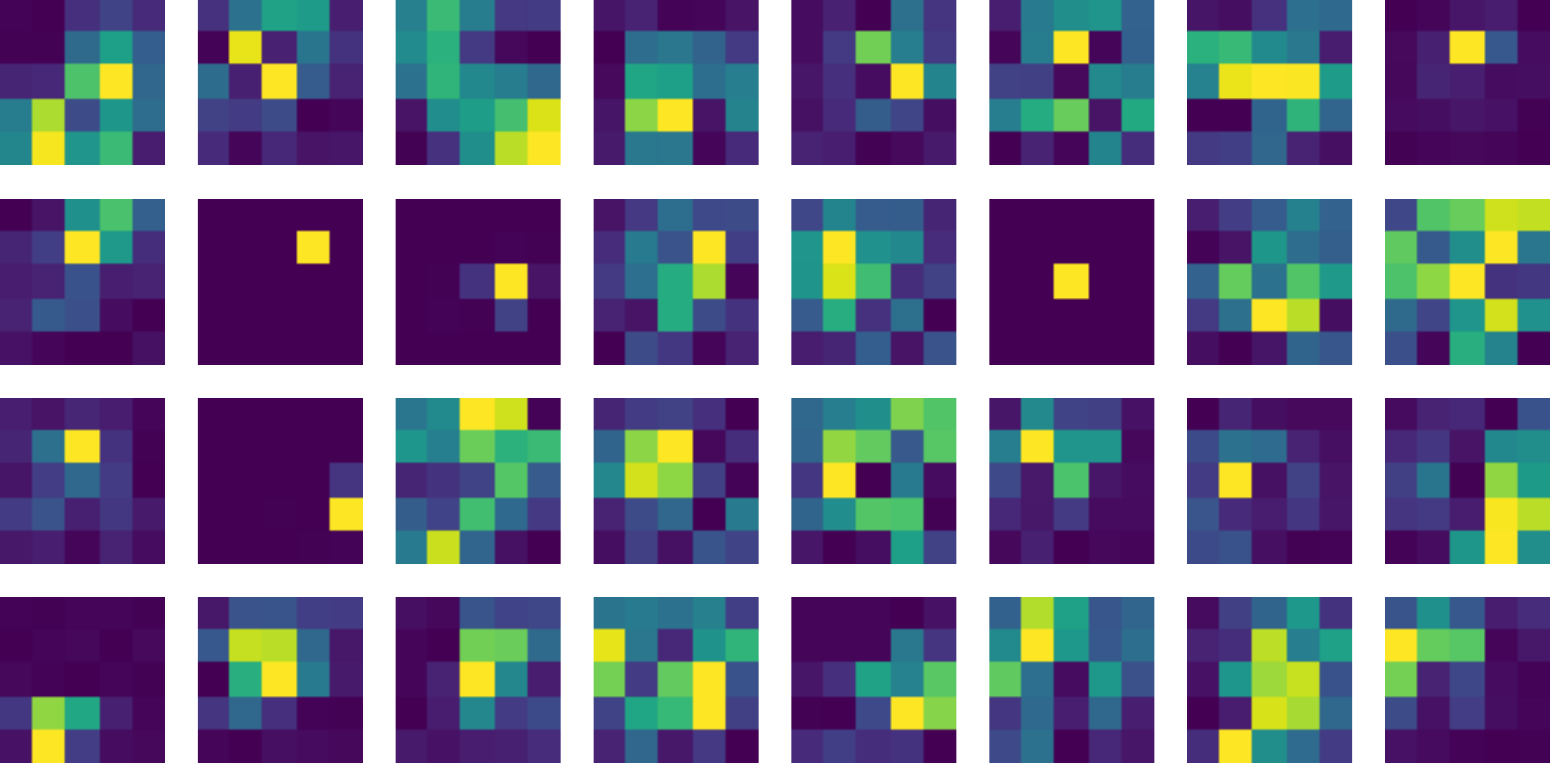}}
  \\
  \includegraphics[width=\columnwidth]{figures/filters/colorbar_1.pdf}
   \caption{A view of each of the (5x5) learned filters of the first layer of \maxt{} and \avgt{} models trained on the MNIST dataset. These models are not susceptible to decision-based attacks as opposed to those in Figures \ref{app:fig:app_filters_3}, \ref{app:fig:app_filters_4}. Notably, we had to employ `ad-hoc' techniques to manipulate the individual perturbation models to be able to train these models. However, even after such manipulations, the accuracy against the worst-case adversary in the union of $\ell_\infty$, $\ell_2$, $\ell_1$ perturbation models for \maxt{}, \avgt{} approaches is considerably worse than the \MSD{} approach.}
  \label{app:fig:app_filters_5}
\end{figure*}

As a result, in order to achieve reasonable performance in case of \maxt{} and \avgt{} models against decision-based attacks, we had to employ methods to manipulate the perturbation models in an `ad-hoc' manner. More specifically, in case of \maxt{} we had to increase the number of restarts of the $\ell_1$ attack during training, and perform an early stop at the end of the fourth epoch (Figure \ref{fig:max_final_filters}) since further training biased the model towards $\ell_\infty$ robustness, and made it susceptible to decision-based attacks. In case of \avgt{}, we had to increase the maximum radius of the $\ell_1$ attack to 12 (Figure \ref{fig:avg_final_filters}). It is worth noting that both the approaches help cosmetically strengthen the relative effect of the $\ell_1$ attack and help reduce the number of sparse filters. We observe that these models perform significantly better against decision-based attacks as opposed to those in Figures \ref{app:fig:app_filters_3}, \ref{app:fig:app_filters_4}.

Finally, we emphasize that while learning sparse convolution filters and the susceptibility to gradient-free attacks is often correlated, there is \textit{no} consistent relation between the ``number'' of such filters and the final model performance or the presence of gradient masking. We perform this empirical analysis for completeness to follow up on previous work by \citet{madry2018towards}, and it comes with no formal statements. In fact, a model may perform better against decision-based attacks even if it has more sparse filters than another model. We hope that these preliminary observations encourage further exploration around the phenomenon of gradient masking in adversarially robust models. 

\section{Attacks outside the perturbation model}
\label{app:outside}
In this section, we present some additional experiments exploring the performance of our model on attacks which lie outside the perturbation model. Note that this is presented only for exploratory reasons and there is no principled reason why the adversarial defenses should generalize beyond the perturbation model defended against. 

\begin{table}
  \caption{Performance on CIFAR-10-C}
  \label{table:cifar10c}
  \centering
  \begin{tabular}{l|r}
    & Accuracy\\
    \hline
    Standard model & 66.0\%\\
    \hline
    $P_\infty$ & 75.0\%\\
    $P_2$ & 82.7\%\\
    $P_1$ & 57.8\%\\
    \hline
    \maxt{} & 70.8\%\\
    \avgt{} & 76.8\%\\
    \MSD{} & 74.2\%\\
    \hline
  \end{tabular}
\end{table}

\paragraph{Common corruptions} We measure the performance of all the models on CIFAR-10-C, which is a CIFAR10 benchmark which has had common corruptions applied to it (e.g. noise, blur, and compression). We report the results in Table \ref{table:cifar10c}. We find that that, apart from the $P_1$ model, the rest achieve some improved robustness against these common corruptions above the standard CIFAR10 model. 

\paragraph{Defending against $\ell_1$ and $\ell_\infty$ and evaluating on $\ell_2$}
We also briefly study what happens when one trains against $\ell_1$ and $\ell_\infty$ perturbation models, while evaluating against the $\ell_2$ adversary. Specifically, we take the \MSD{} approach on MNIST and simply remove the $\ell_2$ adversary from the perturbation model. This results in a model which has its $\ell_1$ and $\ell_\infty$ robust performance against a PGD adversary drop by 1\% and its $\ell_2$ robust performance against a PGD adversary (which it was not trained for) drops by 2\% in comparison to the original \MSD{} approach on all three perturbation models. 

As a result, we empirically observe that including the $\ell_2$ perturbation model in this setting actually improved overall robustness against all three perturbation models. Unsurprisingly, the $\ell_2$ performance drops to some degree, but the model does not lose all of its robustness.


\end{document}